\documentclass[10pt,twocolumn,letterpaper]{article}

\usepackage[pagenumbers]{cvpr} % To force page numbers, e.g. for an arXiv version

\usepackage{graphicx}
\usepackage{amsmath}
\usepackage{amssymb}
\usepackage{booktabs}
\usepackage{multirow}
\usepackage{multicol}
\usepackage{comment}
\usepackage{footnote}
\usepackage{bbding}
\usepackage{pifont}
\usepackage{arydshln}
\newcommand{\cmark}{\ding{51}}%
\newcommand{\xmark}{\ding{55}}%

\newcommand{\mA}{\mathcal{A}}
\newcommand{\mX}{\mathcal{X}}
\newcommand{\mY}{\mathcal{Y}}
\newcommand{\mO}{\mathcal{O}}
\newcommand{\mM}{\mathcal{M}}
\newcommand{\mD}{\mathcal{D}}

\newcommand{\mXt}{\mathcal{X}^{(t)}}
\newcommand{\mYt}{\mathcal{Y}^{(t)}}

\newcommand{\mMt}{\mathcal{M}^{(t)}}

\newcommand{\Pt}{P^{(t)}}
\newcommand{\PtT}{P^{(t)T}}

\usepackage[pagebackref,breaklinks,colorlinks]{hyperref}

\usepackage[capitalize]{cleveref}
\crefname{section}{Sec.}{Secs.}
\Crefname{section}{Section}{Sections}
\Crefname{table}{Table}{Tables}
\crefname{table}{Tab.}{Tabs.}

\def\cvprPaperID{2181} % *** Enter the CVPR Paper ID here
\def\confName{CVPR}
\def\confYear{2023}

\begin{document}

\title{IMP: Iterative Matching and Pose Estimation with Adaptive Pooling}

\author{Fei Xue \quad Ignas Budvytis \quad Roberto Cipolla \\
	{ University of Cambridge}  \\
	{\small \{fx221, ib255, rc10001\}@cam.ac.uk }
}
\maketitle

%%%%%%%%% ABSTRACT
%%%%%%%%% ABSTRACT
\begin{abstract}
	Previous methods solve feature matching and pose estimation using a two-stage process by first finding matches and then estimating the pose. As they ignore the geometric relationships between the two tasks, they focus on either improving the quality of matches or filtering potential outliers, leading to limited efficiency or accuracy. In contrast, we propose an \textbf{i}terative \textbf{m}atching and \textbf{p}ose estimation framework (IMP) leveraging the geometric connections between the two tasks: a few good matches are enough for a roughly accurate pose estimation; a roughly accurate pose can be used to guide the matching by providing geometric constraints. To this end, we implement a geometry-aware recurrent attention-based module which jointly outputs sparse matches and camera poses. Specifically, for each iteration, we first implicitly embed geometric information into the module via a pose-consistency loss, allowing it to predict geometry-aware matches progressively. Second, we introduce an \textbf{e}fficient IMP, called EIMP, to dynamically discard keypoints without potential matches, avoiding redundant updating and significantly reducing the quadratic time complexity of attention computation in transformers. Experiments on YFCC100m, Scannet, and Aachen Day-Night datasets demonstrate that the proposed method outperforms previous approaches in terms of accuracy and efficiency. Code is available at \url{https://github.com/feixue94/imp-release}
	
\end{abstract}
\section{Introduction}
\label{sec:introduction}

\begin{figure}[t]
	\centering
	\includegraphics[width=.98\linewidth]{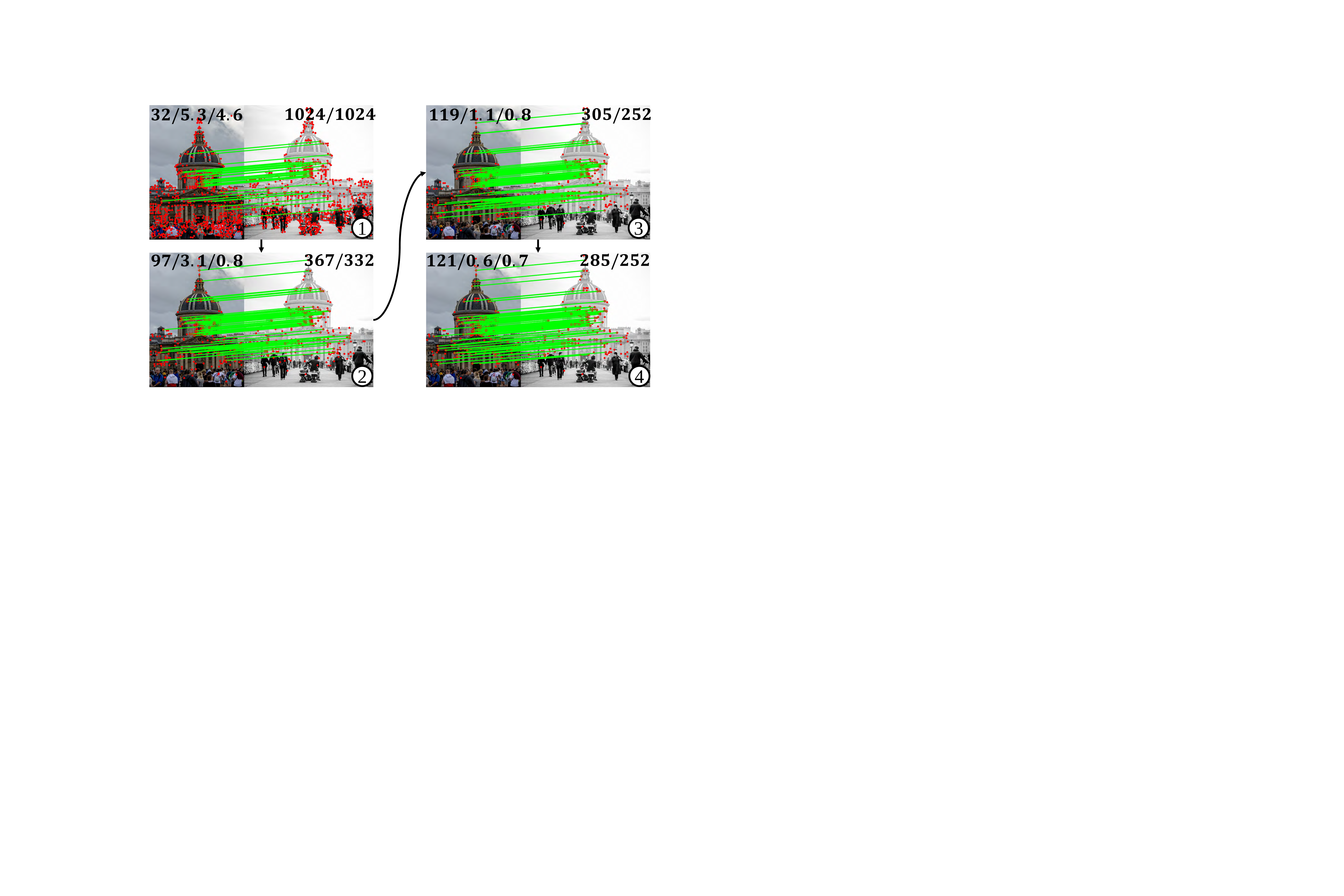}
	%\caption{\textbf{Overview of our framework.} We perform feature matching and pose estimation iteratively. In the iteration process, matches are gradually predicted and estimated pose is getting increasingly more precise. The pose is used to provide geometric guidance to find more matches and discard redundant keypoints, making next iterations faster. The iteration terminates when the pose converges, so as to avoid redundant operations.}
	\caption{\textbf{Process of iterative matching and pose estimation}. For each image pair, we report the number inliers/rotation/translation errors (top-left) and retained keypoints in left/right images (top-right) at iterations from 1 to 4. In the iterative process, our method finds more inliers spanning almost the whole image, estimates increasingly precise pose and discards keypoints without true matches gradually.}
	\vspace*{-10pt}
	\label{fig:overview}
\end{figure}

Feature matching and relative pose estimation are two fundamental tasks in computer vision and especially important to visual localization and 3D reconstruction. Traditionally, the two tasks are performed in two stages separately by first finding correspondences between keypoints extracted from two images with nearest neighbor (NN) matching and then estimating the relative pose from predicted matches with robust estimators, \eg RANSAC~\cite{ransac,usac,magsac,magsac++}. This pipeline has been the de-facto standard framework for decades~\cite{relativepose2022}. However, due to repetitive textures/structures, changing appearances and viewpoint variations, matches given by NN often contain a large number of outliers, leading to poor pose accuracy~\cite{aachen,superglue}. To mitigate this problem, some works~\cite{ce,acne,adalam,oanet,tnet,ms2dg,ngransac,lmcnet} filter potential outliers of predicted matches with neural networks to improve the pose accuracy. Although they report better results, their performance is limited by the quality of initial matches and require extra time for filtering at test time. Alternatively, advanced matchers such as SuperGlue~\cite{superglue} enhance the matching quality directly by using global information from all keypoints via transformers~\cite{attention} with a fixed number (\eg 9) of iterations. These methods have obtained remarkable performance. Yet, their quadratic time complexity for the attention computation degrades the efficiency in real applications. Some following works~\cite{sgmnet,clustergnn,ela2022} explore more efficient variations, they run faster but are significantly less accurate (see Table~\ref{tab:yfcc} and~\ref{tab:aachen}).

In this paper, we aim to introduce an efficient and accurate framework for iterative matching and pose estimation. Our approach is built upon the following observations: (1) a few well distributed matches (\eg 5) could give a roughly accurate pose (\eg essential matrix); (2) in turn, a roughly accurate pose could provide strong geometric constraints (\eg epipolar line) to find more accurate matches at low cost; (3) the pose also reveals which keypoints have potential correspondences, preventing redundant operations. Based on the geometric connections of the two tasks, we propose an \textbf{i}terative \textbf{m}atching and \textbf{p}ose estimation framework (IMP), to perform matching and pose estimation iteratively as opposed to in two separate stages. Specifically, we progressively augment descriptors with self and cross attention as~\cite{superglue,sgmnet,clustergnn,ela2022}, find matches and estimate the relative pose. As descriptors get gradually more discriminative, more correct matches can be found, leading to increasingly more precise pose, as shown in Fig.~\ref{fig:overview}. However, due to the noise~\cite{pixelperfect} and degeneration (\eg co-planar keypoints)~\cite{degensac}, not all inliers could give a good pose~\cite{goodmodel,instability}. In addition to the classification loss mainly used by prior methods~\cite{superglue,sgmnet}, we apply a pose-consistency loss~\cite{oanet} to the matching process, enabling the model to find matches which are not only accurate but also able to give a good pose.

Moreover, in order to avoid redundant operations on uninformative keypoints, we employ a sampling strategy by combining the matching and attention scores of keypoints and the uncertainty of predicted poses to adaptively remove useful keypoints, as shown in Fig.~\ref{fig:overview}. Compared with prior sampling approaches~\cite{deit,adavit2022} based mainly on attention scores, our adaptive strategy overcomes the over-sampling problem effectively. Our framework reduces the time cost from two aspects. First, in contrast to adopting a fixed number of iterations for all cases~\cite{superglue,sgmnet,clustergnn}, it runs fewer iterations for easy cases with few viewpoint or appearance changes and more for challenging cases. Second, it reduces the cost of each iteration, significantly reducing the quadratic time complexity of attention computation. We also show that discarding potential outliers increases not only efficiency but also accuracy (see Sec.~\ref{sec:experiments}). The efficient version of IMP is called EIMP. Ours contributions are as follows:

% although a large number of pooling strategies~\cite{adavit2022,deit} are proposed to reduce the complexity of transformers, they leverage purely the attention scores. Directly applying these techniques to the matching task leads to over-pruning. Alternatively, 
%When a large number of outliers are removed, our method gives higher efficiency (see Sec.~\ref{sec:experiments}).

%However, several challenges exist. Firstly, due to the noise~\cite{pixelperfect} and degeneration of keypoints' locations (\eg co-plane)~\cite{degensac}, not all inliers could give a good pose~\cite{goodmodel,instability}. Previous matching methods~\cite{superglue, sgmnet,clustergnn} such as SuperGlue focus mainly on predicting correct matches with a classification loss for training but ignore the geometric requirements of matches. Instead, we additionally apply a geometric loss~\cite{oanet} to transformers, enabling the model to give matches which are not only accurate but able to give a good pose. Secondly, discarding keypoints based only on attention scores is prone to over-pruning, degrading the accuracy, especially at early iterations when attention scores are not informative. Instead,

\begin{itemize}
	\item We propose to perform geometry-aware matching and pose estimation iteratively, allowing the two tasks to boost each other in an iterative manner.
	
	\item We adopt a robust sampling strategy to adaptively discard redundant keypoints in the iteration process, significantly decreasing the time complexity. 
	
	\item We apply the pose uncertainty to the sampling strategy, which further improves the accuracy matching and pose estimation.
	
\end{itemize}

Our experiments on relative pose estimation and large-scale localization tasks demonstrate that our method outperforms previous competitors and is more efficient. We organize the rest of the paper as follows. In Sec.~\ref{sec:related_works}, we discuss related works. In Sec.~\ref{sec:method:imp}, we give a detailed description of our method. We test the performance of our model in Sec.~\ref{sec:experiments} and conclude the paper in Sec.~\ref{sec:conclusion}.

\section{Related works}
\label{sec:related_works}
In this section, we discuss related work on local feature matching, efficient attention, and outlier filtering.

\begin{figure*}[t]
	\centering
	\begin{subfigure}{.56\textwidth}
		\centering
		\includegraphics[width=.99\linewidth]{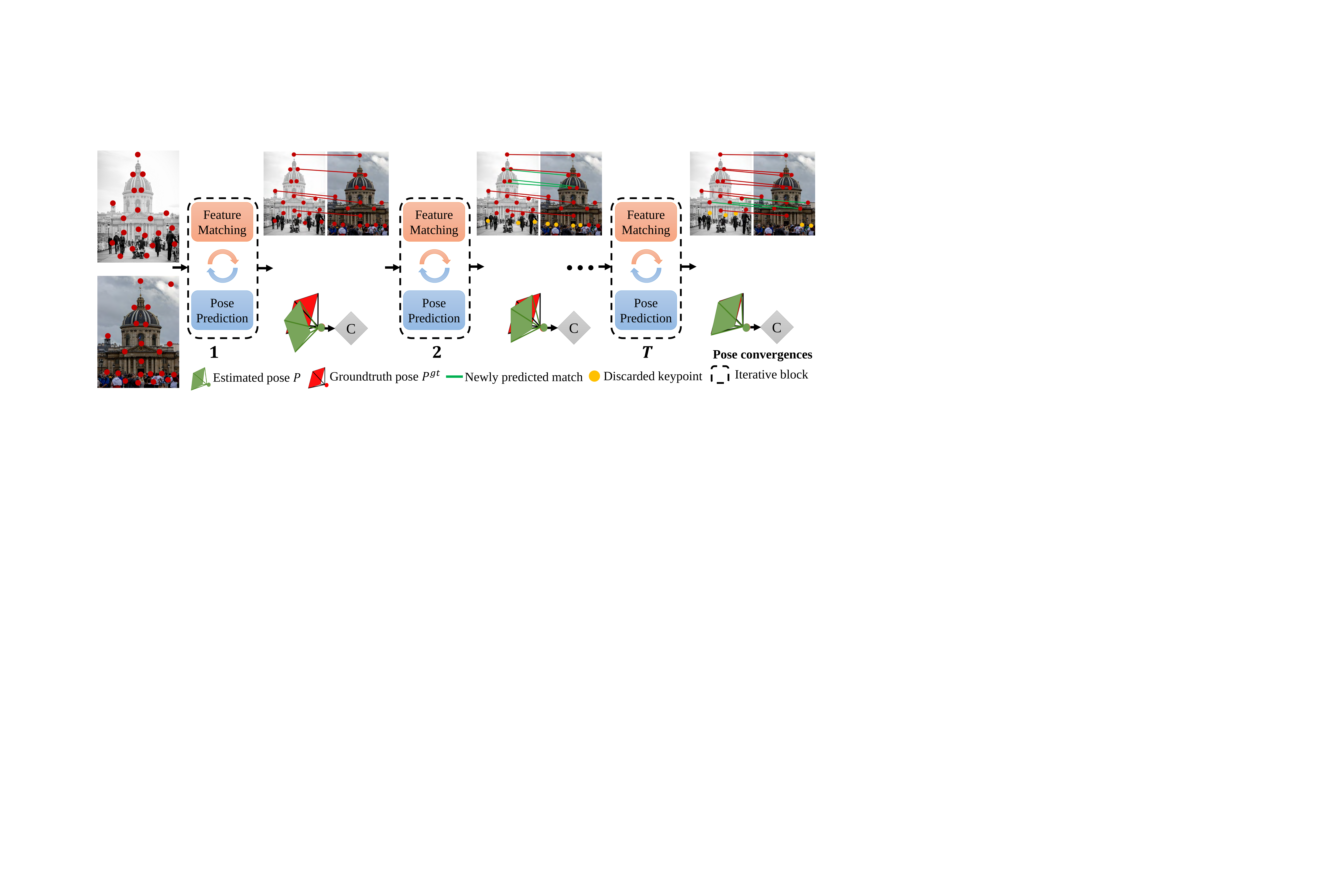}
		\caption{Pipeline of our method}
		\label{fig:pipeline}
	\end{subfigure}%
	\begin{subfigure}{.25\textwidth}
		\centering
		\includegraphics[width=.97\linewidth]{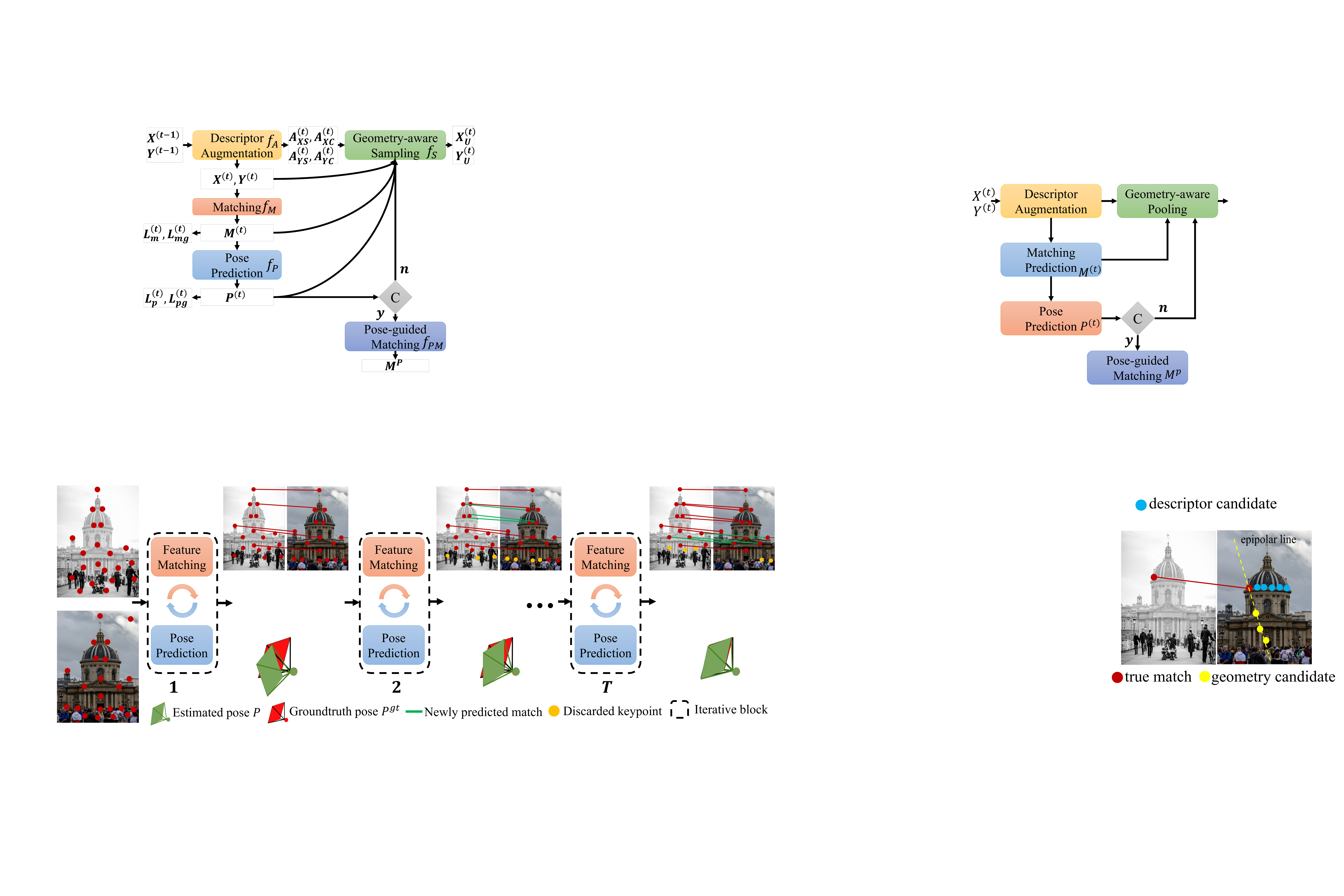}
		\caption{$t$th iterative block}
		\label{fig:block}
	\end{subfigure}%
	\begin{subfigure}{.19\textwidth}
		\centering
		\includegraphics[width=.98\linewidth]{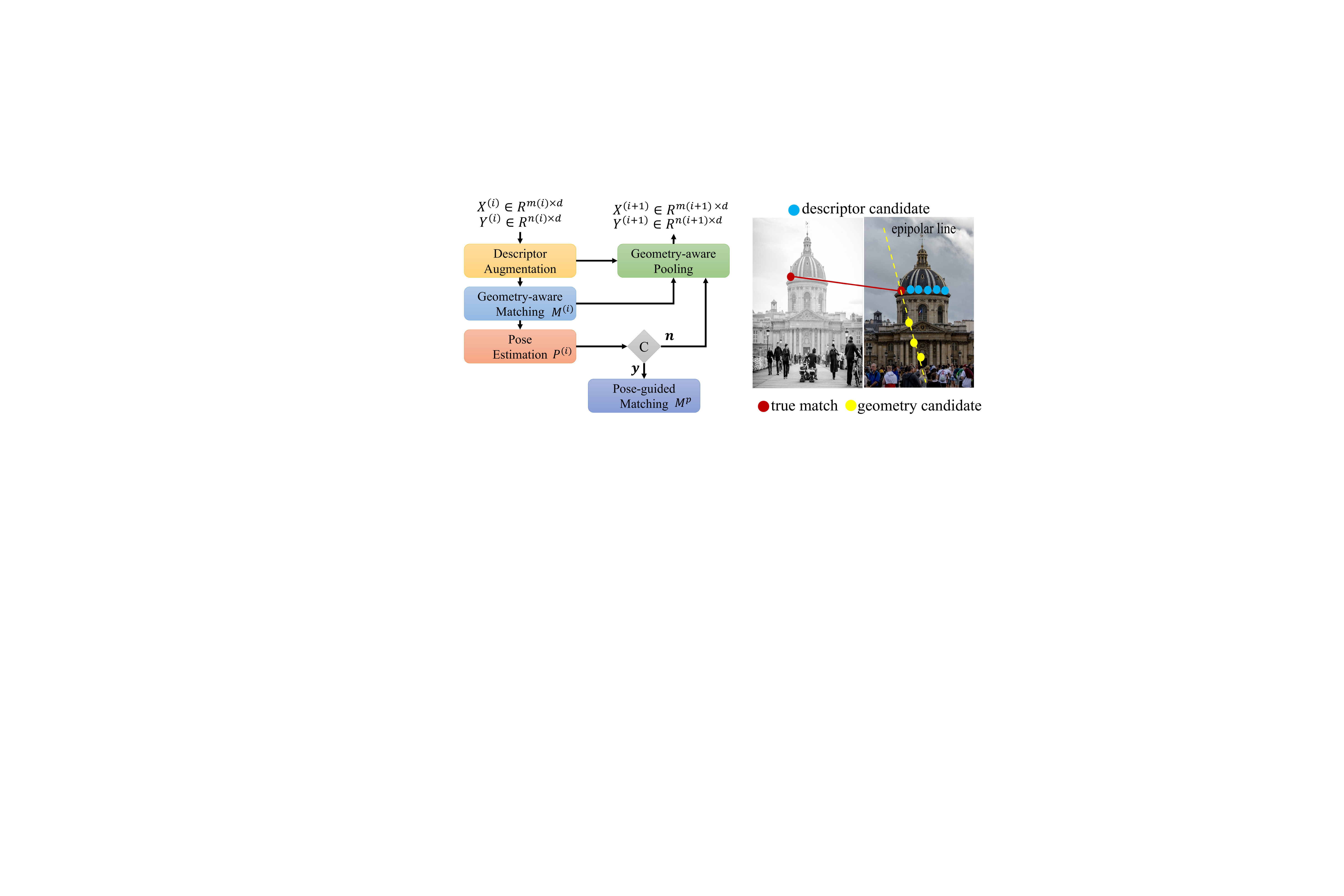}
		\caption{Pose-guided matching}
		\label{fig:pose_guided_matching}
	\end{subfigure}
	%\vspace*{-8pt}
	\caption{\textbf{Pipeline of our method.} In the iteration process, descriptors are gradually augmented by our recurrent attention-based module. Augmented descriptors are then used to compute matches $\mM^{(t)}$, which are further used for estimating the pose $P^{(t)}$. As descriptors become more discriminative, more correct matches can be found, leading to more precise poses. The pose is utilized to provide geometric guidance to find more matches and discard redundant keypoints with geometry-aware sampling $f_{S}$, making next iterations faster. If the pose converges, iteration stops. Our recurrent attention-based module at $t$th iteration (b) and pose-guided matching (c) are also visualized.}
	\vspace*{-10pt}
	%\label{fig:pipeline}
\end{figure*}

\textbf{Local feature matching.} Traditionally, NN matching and its variants, \eg, mutual NN (MNN) and NN with ratio test (NN-RT)~\cite{sift} are widely used for finding correspondences between two sets of keypoints~\cite{sift, orb, superpoint, rootsift2012, sfd2}. Since they perform point-wise matching, they are fast but not robust to large viewpoint and appearance changes, \eg illumination and season variations. Recently, SuperGlue~\cite{superglue} applies graph connections~\cite{glnet,lsg} to keypoints and adopts with transformers~\cite{attention} with self and cross attention to embed spatial information and achieves remarkable performance.
Despite its excellent accuracy, its limitations are twofold. First, the complexity of attention is quadratic to the number of keypoints, degrading the efficiency in real applications. Second, it adopts a fixed number (9) of layers for message propagation for all input cases. This causes extra time cost especially for simple cases with few viewpoint and appearance changes, which require much fewer number of iterations to find potential inliers. Therefore, running a fixed number iterations for all cases leads to redundancy.

Some variants~\cite{sgmnet,clustergnn,ela2022} try to solve the first problem by using a fixed number of seeded keypoints for message propagation~\cite{sgmnet}, performing cluster-wise matching~\cite{clustergnn} or aggregating local and global information separately~\cite{ela2022}. In spite of their high efficiency, they lose significant accuracy. Besides, like SuperGlue, they also update keypoints without correspondences redundantly in each iteration and adopt a fixed number of iterations for all cases constantly. Unlike these approaches, our model utilizes predicted poses to prevent extra iterations for easy cases. Additionally, as our method removes keypoints without true matches adaptively to reduce time complexity in each iteration, both efficiency and accuracy are guaranteed. 

\textbf{Efficient attention.} Many works are proposed to mitigate the quadratic time complexity of the attention mechanism~\cite{attention}. They reduce the complexity by learning a linear projection function~\cite{linearattention,linformer2020}, a token selection classifier~\cite{dynamicvit,adavit2022,setformer2019} or shared attention~\cite{shareattention}, to name a few. We refer the reader a comprehensive survey~\cite{efficient_transformer} for more details. Since most of these methods are designed to extract high-level features for downstream tasks \eg image recognition~\cite{deit,adavit2022,shareattention,linearattention,dino2021,dinofeat2022} usually with a fixed number of tokens, directly transferring these techniques to the matching task which has dynamic numbers of keypoints as input could cause performance loss. Instead, we utilize the geometric properties of the matching task to adaptively discard redundant keypoints. Our strategy depends mainly on the number of inliers in the input and is completely adaptive.

\textbf{Outlier filtering.} Ratio test~\cite{sift} is often used to remove matches with high uncertainties. Recently, CNNs are used to leverage the spatial context of input keypoints to filter potential outliers~\cite{ce,acne,adalam,oanet,tnet,ms2dg,ngransac,lmcnet}. ACNe~\cite{acne} introduces an attention mechanism to gather information from other keypoints for filtering. OANet~\cite{oanet} embeds geometric priors to further improve the performance. LMCNet~\cite{lmcnet} and CLNet~\cite{clnet} use local motion consistency to filter inconsistent matches. MS2DG-Net~\cite{ms2dg} infuses semantics into a graph network to enhance the accuracy. These methods have achieved better results than ratio test and could serve as an additional module to filter potential outliers. However, their performance relies heavily on the quality of initial matches and require extra cost for filtering at test time. In contrast, our method embed geometric information directly into the matching module, allowing the model to predict accurate and pose-ware correspondences end-to-end.

\begin{figure*}[t]
	\centering
	\includegraphics[width=.99\linewidth]{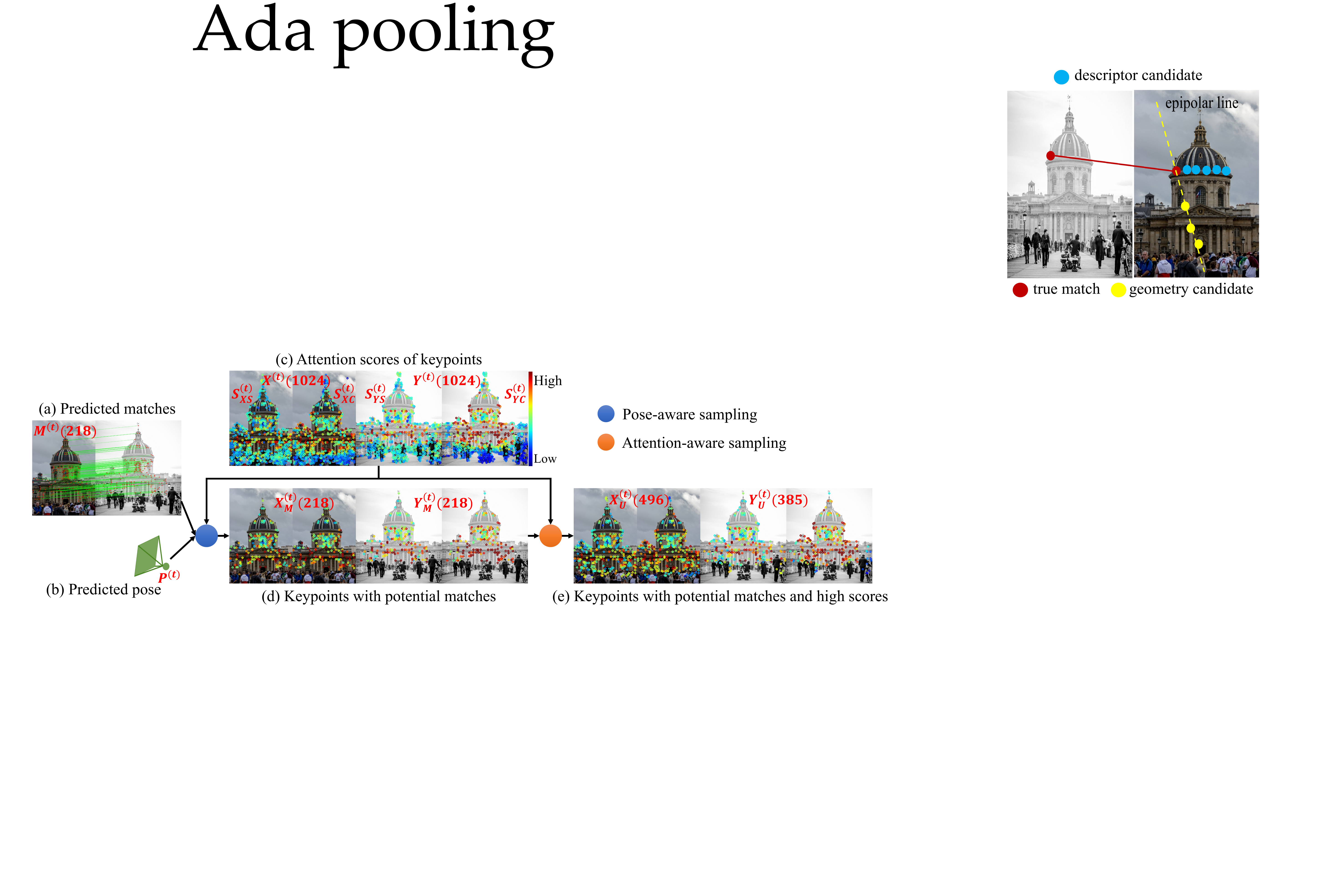}
	%\vspace*{-8pt}
	\caption{\textbf{Adaptive sampling.} At iteration $t$, predicted matches $\mM^{(t)}$ and pose $P^{(t)}$ are used to select keypoints with potential correspondences. These selected keypoints are expanded by those with high self or cross attention scores. The finally preserved keypoints are those with potential matches and high contribution, guaranteeing both the efficiency and accuracy.}
	\vspace*{-10pt}
	\label{fig:ada_pooling}
\end{figure*}
\section{Method}
\label{sec:method:imp}
We first give an introduction to performing iterative transformer-based matching in Sec.~\ref{sec:method:matching}. Then, we describe our iterative pose estimation framework, the adaptive sampling strategy and application at test time in Sec.~\ref{sec:pose_matching}, Sec.~\ref{sec:dyn_attn} and Sec~\ref{sec:test_time}, respectively. We visualize the pipeline of our framework in Fig.~\ref{fig:pipeline}.

\subsection{Iterative transformer-based matching}
\label{sec:method:matching}
In this section, we introduce how to perform transformer-based matching iteratively.

\textbf{Problem formulation.} Given two sets of keypoints (\eg SuperPoint~\cite{superpoint} or SIFT~\cite{sift}) $\mX=\{x_1, x_2,...,x_m\}, \mathcal{Y}=\{y_1, y_2,...,y_n\}$ ($m$, $n$ are the number of keypoints) extracted from two images, matchers predict matches between $\mathcal{X}$ and $\mathcal{Y}$, denoted as $\mO=\{o_1, o_2, ..., o_k\}$, where $o_j=(x_k,y_l)$ is the matched pair. Each keypoint $x_i=(u_i, v_i,c_i,\mathbf{d}_i)$ comprises its 2D coordinates $(u_i,v_i)$, confidence $c_i$ and descriptor $\mathbf{d}_i\in R^{d}$ ($d$ is the descriptor size). 

\textbf{Descriptor augmentation.} As~\cite{superglue,sgmnet,clustergnn}, for each keypoint $x_i$, its coordinates $(u_i,v_i)$ and confidence $c_i$ are encoded into a high-dimension vector with a multi-layer perception (MLP) $f_{enc}$, which is then added to its descriptor $\mathbf{d}_i$, as: $\mathbf{d}_i^{'} = \mathbf{d}_i + f_{enc}(u_i,v_i,c_i)$. $\mathbf{d}_i^{'}$ is used to replace $\mathbf{d}_i$ as input for descriptor augmentation, as: 

\begin{footnotesize}
	\vspace*{-10pt}
\begin{align}
	\mA_{XS}^{(t)} = softmax(\frac{f^{S}_q(\mX^{(t-1)})(f^{S}_k(\mX^{(t-1)}))^{T}}{\sqrt{d}}),\\
	\mA_{XC}^{(t)} = softmax(\frac{f^{C}_q(\mX^{(t-1)})(f^{C}_k(\mY^{(t-1)}))^{T}}{\sqrt{d}}), \\ \nonumber
	\mX^{(t)} = \mX^{(t-1)} + f_{mlp}^{S}(f_{p}^{S}(\mA_{XS}^{(t)} (f^{S}_v(\mX^{(t-1)}))) ||  \mX^{(t-1)}) \\  + f_{mlp}^{C}(f_{p}^{C}(\mA_{XC}^{(t)} (f^{C}_v(\mY^{(t-1)}))) || \mX^{(t-1)}). 
\end{align}
\vspace*{-12pt}
\end{footnotesize}

%\begin{align}
%	\mX^{(t)},\mA^{(t)}_{XS},\mA^{(t)}_{XC},\mY^{(t)},\mA^{(t)}_{YS}, \mA^{(t)}_{YC} = f_A(\mX^{(t-1)},\mY^{(t-1)}). \nonumber
%\end{align}
%
$\mA_{XS}^{(t)}$ and $\mA_{XC}^{(t)}$ are self (S) and cross (C) attention matrices for $\mX^{(t)}$. $f^{S/C}_{q/k/v/p}$ are FC layers. $d$ is descriptor dimension. $f_{mlp}^{S/C}$ are 3-layer MLPs. $||$ is channel-wise concatenation. We use sharing attention mechanism~\cite{shareattention} to further augment descriptors with pre-computed attention matrices:

\begin{footnotesize}
	\vspace*{-10pt}
\begin{align}
	\nonumber
	\mX^{(t)} = \mX^{(t)} + f_{mlp}^{S}(f_{p}^{S}(\mA_{XS}^{(t)} (\bar{f}^{S}_v(\mX^{(t-1)}))) ||  \mX^{(t-1)}) \\  + f_{mlp}^{C}(\phi_{p}^{C}(\mA_{XC}^{(t)} (\bar{f}^{C}_v(\mY^{(t-1)}))) || \mX^{(t-1)}).
\end{align}
\vspace*{-12pt}
\end{footnotesize}

$\bar{f}^{S/C}_{v}$ are FC layers as well. We conduct the same operations to $\mY^{(t-1)}$ to obtain augmented descriptors $\mY^{(t)}$, self $\mA_{YS}^{(t)}$ and cross $\mA_{YC}^{(t)}$ attention matrices. As shown in Fig.~\ref{fig:block}, $\mX^{(t)}$ and $\mY^{(t)}$ are augmented descriptors used for the next iteration. $\mA^{(t)}_{XS}$, $\mA^{(t)}_{XC}$, $\mA^{(t)}_{YS}$ and $\mA^{(t)}_{YC}$ are self and cross attention matrices used for message propagation in the augmentation function $f_A$. In prior works~\cite{superglue,sgmnet}, self and cross attention matrices are only used for augmenting descriptors. However, in our model, they are further utilized for adaptive sampling (Sec.~\ref{sec:dyn_attn}). 

%Unlike the standard attention used by SuperGlue~\cite{superglue}, we adopt sharing attention mechanism which has been proved effective for image recognition~\cite{shareattention}. In detail, $f_A$ propagates information twice with a shared attention matrix computed only once, allowing keypoints to gather more message at low cost. In prior methods~\cite{superglue,sgmnet}, self and cross attention matrices are only used for augmenting descriptors. However, in our model, they are further utilized for adaptive sampling (see Sec.~\ref{sec:dyn_attn}).

\textbf{Iterative matching prediction.} Augmented descriptors $\mX^{(t)}, \mY^{(t)}$ are used to compute the matching matrix $\mM^{(t)} \in R^{m \times n}$ with function $f_M$. $f_M$ first computes Euclidean distance $\mD^{(t)} \in R^{m \times n}$ for all pairs in $\mX^{(t)}_A \times \mY^{(t)}_A$ and then uses Sinkhorn algorithm~\cite{sinkhorn1967,sinkhorn2013} to obtain $\mM^{(t)}$ by optimizing $\mD^{(t)}$ a transport problem~\cite{transport2017}. Matches with scores over than a certain threshold $\theta_m$ are deemed as predicted matches. In our framework, we predict matches at each iteration $t$ for pose estimation and keypoint sampling as opposed to at only the last iteration~\cite{superglue,clustergnn}.

\textbf{Iterative matching loss.} Similar to SuperGlue~\cite{superglue}, we adopt the classification loss of minimizing the negative log-likelihood of the matching matrix to enforce the network to predict correct matches for each iteration $t$, as: 

\begin{footnotesize}
	\vspace*{-10pt}
\begin{align}
	\label{eq:cls}
	\nonumber
	L^{(t)}_{m} &=-\sum_{(i,j) \in \mM^{gt}}log\bar{\mM}^{(t)}_{i,j} \\ 
		& - \sum_{i \in \bar{\mM}^{gt}}log\bar{\mM}^{(t)}_{i,n+1} -\sum_{j\in \bar{\mM}^{gt}}log\bar{\mM}^{(t)}_{m+1,j}.
\end{align}
\vspace*{-8pt}
\end{footnotesize}

$\bar{\mathcal{M}}^{(t)}$ is the expanded matrix of $\mM^{(t)}$ with an additional row and column for unmatched keypoints~\cite{superglue} in $\mX^{(t)}$ and $\mY^{(t)}$. $\mM^{gt}$ and $\bar{\mM}^{gt}$ are the groundtruth matching matrix and its expansion. Applying the matching loss to each iteration $t$ enables our model to predict correct matches at each iteration progressively.

\subsection{Iterative pose estimation}
\label{sec:pose_matching}
%As shown in Fig.~\ref{fig:block}, in each iteration $t$, descriptors are first augmented by function $f_A$.  Then augmented descriptors are used to find matches $\mM^{(t)}$ with function $f_M$. Next, predicted matches are utilized for estimating the pose $P^{(t)}$ with estimator $f_P$. The sampling module $f_S$ accepts augmented descriptors, matches $\mM^{(t)}$ and pose $P^{(t)}$ as input, and adaptively discard redundant keypoints. If the pose converges, we perform pose-guided matching $f_{PM}$ to find more matches $M^P$.
In this section, we give a detailed description of how to predict poses and the usage of predicted poses for enhancing the matching process in each iteration, as shown in Fig.~\ref{fig:block}.

\textbf{Pose-consistency loss.} Because of the noise~\cite{pixelperfect} and degenerated keypoints~\cite{degensac}, not all correct matches could give a good
pose~\cite{goodmodel,instability}. Eq (\ref{eq:cls}) only guarantees that keypoints with more discriminative descriptors have higher matching scores and are recognized first to attend pose estimation, but ignore the geometric requirements needed for robust pose estimation. Therefore, directly using all potential inliers with matching score over than $\theta_m$ for pose estimation could be inaccurate. Instead, we implicitly infuse geometric information into matching transformers, enforcing the matching module to focus first on matches which are not only correct but also have high probability to give a good pose.

To this end, at each iteration $t$, matches with score over $\theta_m$ in $\mM^{(t)}$ along with their scores are used for fundamental matrix estimation with weighted 8 points function $f_{P}$, as:

\begin{footnotesize}
	\vspace*{-10pt}
\begin{align}
	P^{(t)} = f_{P}(\{(x_i^{(t)},y_i^{(t)},s_i^{(t)})\}).
\end{align}
\vspace*{-12pt}
\end{footnotesize}

$P^{(t)}\in R^{3\times3}$ is the predicted fundamental matrix and $\{(x_i^{(t)},y_i^{(t)},s_i^{(t)})\}$ are predicted matches with matching score $s_i^{(t)}=M^{(t)}[idx(x_i^{(t)}),idx(y_i^{(t)})]$ ($idx(.)$ is the index function). We enforce the geometric consistency between $P^{(t)}$ and the groundtruth $P^{gt}$ by minimizing the pose and epipolar errors jointly, as:

\begin{footnotesize}
	\vspace*{-10pt}
\begin{align}
	L^{(t)}_{p} &= ||P^{(t)}-P^{gt}||_2, \\
	L^{(t)}_{pg} &= \frac{1}{N_{gt}} \sum_{k} f_{epipolar}(P^{(t)}, x^{gt}_k, y^{gt}_k), \\
	L^{(t)}_{mg} &= \frac{1}{N^{(t)}_{p}}\sum_{k} f_{epipolar}(P^{gt}, x^{(t)}_k, y^{(t)}_k).
\end{align}
\vspace*{-12pt}
\end{footnotesize}

$f_{epipolar}$ is Sampson distance~\cite{mvg}.  $(x^{gt}_k, y^{gt}_k)$ and $(x^{(t)}_k, y^{(t)}_k)$ are the groundtruth and predicted matches. $L^{(t)}_p$ and $L^{(t)}_{pg}$ enforce the predicted pose $P^{(t)}$ to be correct with constraints provided by the groundtruth pose $P^{gt}$ and matches $(x^{gt}_k, y^{gt}_k)$, respectively. $L^{(t)}_{mg}$ additionally ensures the correctness of predicted matches $(x^{(t)}_k, y^{(t)}_k)$ with groundtruth pose $P^{gt}$. $N_{gt}$ and $N^{(t)}_{p}$ are the number of groundtruth and predicted matches.

For each iteration, our final loss combines $L^{(t)}_{m}$, $L^{(t)}_{pg}$ and $L^{(t)}_{mg}$ with weights of $\alpha_{m}, \alpha_{p}$ and $\alpha_{g}$, as:

\begin{footnotesize}
	\vspace*{-10pt}
\begin{align}
	L^{(t)} = \alpha_{m}L^{(t)}_{m} + \alpha_{p}L^{(t)}_{p} + \alpha_{g}(L^{(t)}_{pg} + L^{(t)}_{mg}).
\end{align}
\vspace*{-12pt}
\end{footnotesize}

We apply $L^{(t)}$ to each iteration and compute the total loss over $T$ iterations, as $L_{total} = \frac{1}{T}\sum_{t}^{T}L^{(t)}$.

\subsection{Adaptive geometry-aware sampling}
\label{sec:dyn_attn}

In fact, lots of keypoints are uninformative and a large number of keypoints don't have correspondences in the other image (about 70\% of 2k keypoints in YFCC100m dataset~\cite{yfcc100m}). Updating these keypoints leads to extra time cost, so we propose an effective strategy to remove these keypoints, as shown in Fig.\ref{fig:ada_pooling}.

\textbf{Informative keypoints.} The information contained by each keypoint is defined by its contribution to others in the attention matrix $\mA^{(t)}_{XS} \in R^{m \times n \times h}$ ($m, n$ are the number of keypoints in the query and key and $h$ is the number of heads). As~\cite{efficient_transformer,deit}, we compute the score of each keypoint by averaging values along the head and key dimension, as $S^{(t)}_{XS} = \frac{1}{n*h}\sum_{(i,j)}\mA_{XS.,i,j}.$ $S_{XS}\in R^{m}$ is normalized with sum of 1. We also compute scores from $\mA^{(t)}_{XC}$, $\mA^{(t)}_{YS}$, $\mA^{(t)}_{YC}$ as $S^{(t)}_{XC}$, $S^{(t)}_{YS}$, $S^{(t)}_{YC}$. Self and cross attention scores of 1024 keypoints in $\mX^{(t)}$ and $\mY^{(t)}$ are visualized in Fig.~\ref{fig:ada_pooling}c. Although attention scores can be directly used for sampling keypoints by keeping a certain ratio of keypoints with the highest scores~\cite{deit,setformer2019}, this strategy is prone to over-pruning. Instead, we use keypoints with potential matches as guidance to mitigate this problem. 
% At iteration $t$, $\mA$ is one of $\mA^{(t)}_{XS}$ ,$\mA^{(t)}_{XC}$, $\mA^{(t)}_{YS}$, $\mA^{(t)}_{YC}$

%\textbf{Keypoints with potential matches.} 
%  We observe that keypoints with potential correspondences usually have high attention scores.

\begin{figure}[t]
	\centering
	\begin{subfigure}{.22\textwidth}
		\centering
		\includegraphics[width=.95\linewidth]{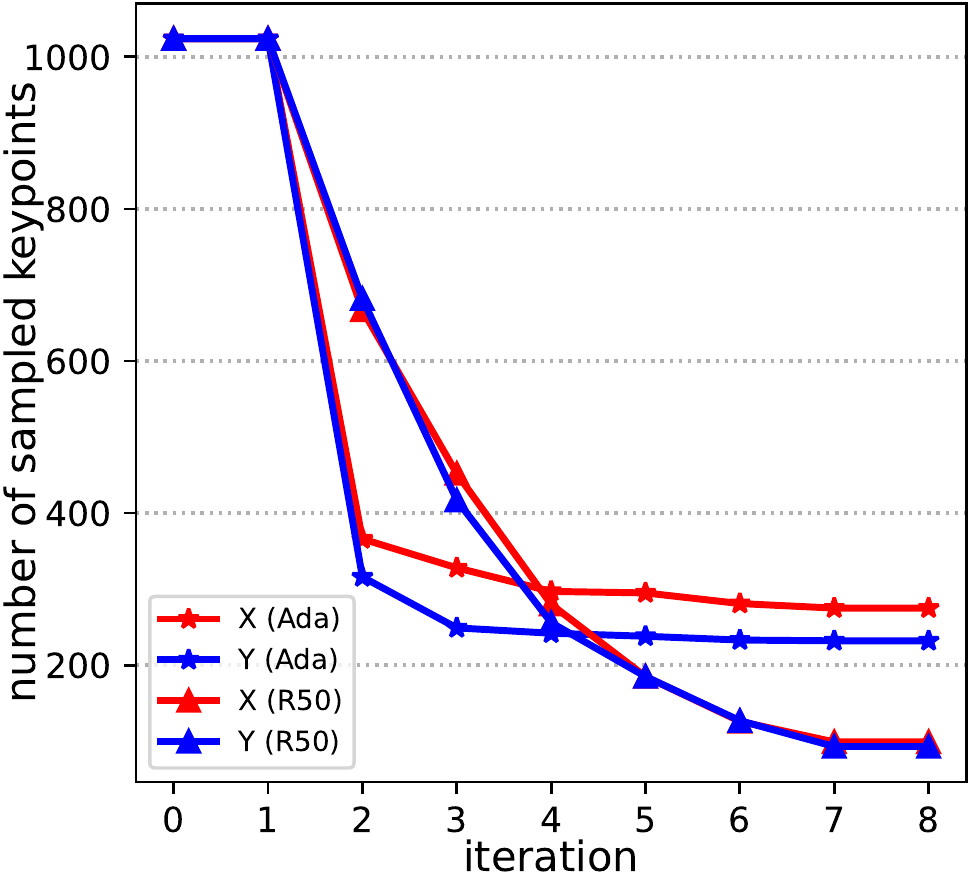}
		\caption{}
		\label{fig:num_sample}
	\end{subfigure}\quad
	\begin{subfigure}{.22\textwidth}
		\centering
		\includegraphics[width=.95\linewidth]{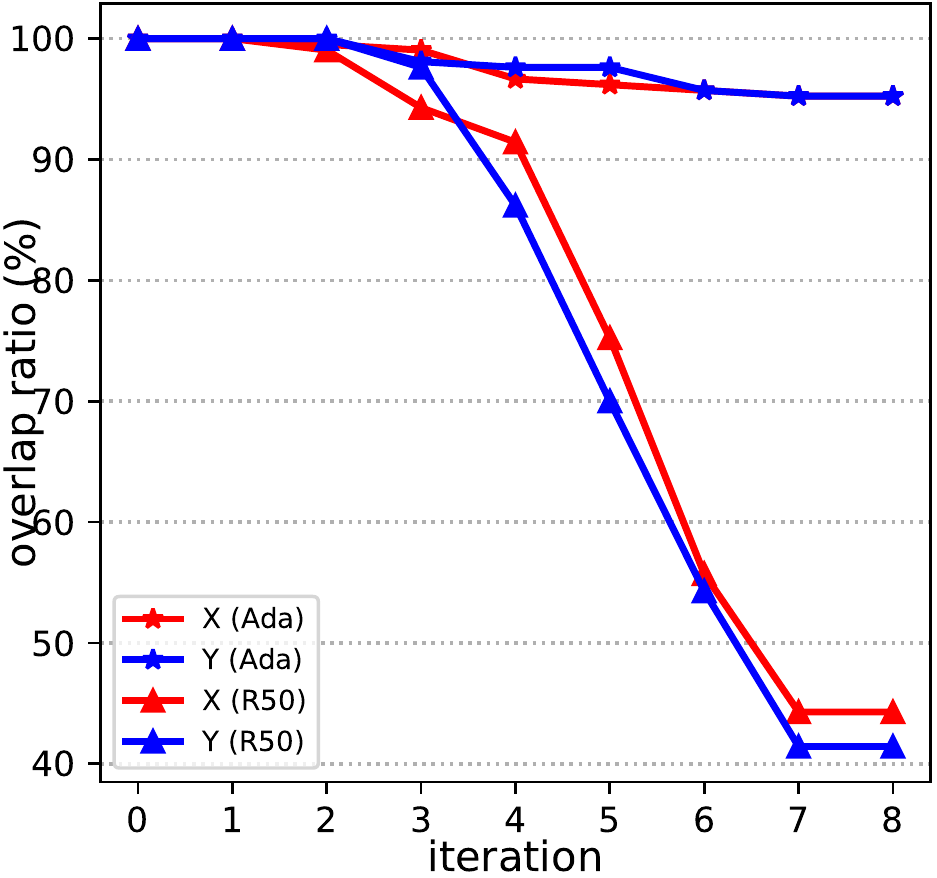}
		\caption{}
		\label{fig:overlap_sample}
	\end{subfigure}
	\vspace*{-10pt}
	\caption{\textbf{Number of Retained keypoints and overlap ratio.} We show (a) the number of retained keypoints and (b) ratio of inliers in two sets at each iteration. Compared to R50, the adaptive sampling reduces redundant keypoints more effectively at early stages (a) and preserves much more inliers at latter iterations (b).}
	\label{fig:num_overlap_sample}
	\vspace*{-10pt}
\end{figure}

\textbf{Adaptive sampling.} At iteration $t$, the matching matrix $\mM^{(t)}$ containing matching confidence of all pairs, reveals which keypoints potentially have true correspondences. We visualize the potential matches and scores of self and cross attention in Fig.~\ref{fig:ada_pooling}a and Fig.~\ref{fig:ada_pooling}d.

Based on matching matrix $\mM^{(t)}$, we generate two subsets $\mX_\mM^{(t)} \subseteq \mX^{(t)}$ and $\mY_\mM^{(t)} \subseteq \mY^{(t)}$ which contain matched keypoints (matching score over $\theta_m$). Since keypoints in $\mX_\mM^{(t)}$ and $\mY_\mM^{(t)}$ are potential inliers, they can provide guidance to find more informative ones. In detail, let $S^{(t)}_{MSX}=\{S^{(t)}_{XS}[idx(x)], s.t.\quad x\in\mX_\mM^{(t)}\}$ be self-attention scores of keypoints in $\mX_\mM^{(t)}$. We generate another set of keypoints with high self-attention scores as $\mX_{S}^{(t)} = \{x, s.t. \quad x \in \mX^{(t)}, S^{(t)}_{XS}[idx(x)] \geq f_{md}(S^{(t)}_{MSX}) \}$ ($f_{md}$ returns the median value). By repeating this process, we obtain another subset from $\mX^{(t)}$ with high cross-attention scores as $\mX_{C}^{(t)}$ and two sets $\mY_{S}^{(t)}$, $\mY_{C}^{(t)}$ from $\mY^{(t)}$ with high self and cross attention scores as well. The final sets are the union of informative keypoints and those with matches, as $\mX_{U}^{(t)}=\mX_\mM^{(t)} \cup \mX_{S}^{(t)} \cup \mX_{C}^{(t)}, \mY_{U}^{(t)}=\mY_\mM^{(t)} \cup \mY_{S}^{(t)} \cup \mY_{C}^{(t)}$. As shown in Fig.~\ref{fig:ada_pooling}e, $\mX_{U}^{(t)}, \mY_{U}^{(t)}$ will replace $\mX^{(t)},\mY^{(t)}$ to join the next iteration. The sampling reduces the number of keypoints from 1024 to 496 and 358 in two sets, reducing the time cost significantly for next iterations. 

Fig.~\ref{fig:num_sample} visualizes the number of keypoints in $\mX_{U}^{(t)}$ and $\mY_{U}^{(t)}$. Fig.~\ref{fig:overlap_sample} shows the value of $\frac{|\mX_{U}^{(t)}\cap X_F|}{|X_F|}$ and $\frac{|\mY_{U}^{(t)}\cap Y_F|}{|Y_F|}$ ($X_F$ and $Y_F$ are two sets of matched keypoints without any sampling; $|.|$ indicates the number of elements), which measure the ability of retaining inliers. For comparison, we also generate two sets of keypoints based on sampling ratio by choosing the top 50\% keypoints with highest self and cross attention scores as $\mX_{R50S}^{(t)}$, $\mX_{R50C}^{(t)}$, $\mY_{R50S}^{(t)}$, and $\mY_{R50C}^{(t)}$. The final sampled sets are $\mX_{R50}^{(t)}=\mX_{R50S}^{(t)} \cup\mX_{R50C}^{(t)}$ and $\mY_{R50}^{(t)}=\mY_{R50S}^{(t)} \cup\mY_{R50C}^{(t)}$. The number of retained keypoints and overlap ratios are also visualized.

Fig.~\ref{fig:num_sample} shows the adaptive strategy performs more effectively at discarding keypoints at early stages and preserving keypoints at later iterations. Fig.~\ref{fig:overlap_sample} illustrates that although R50 has the close ratio to the adaptive strategy at the 2nd and 3rd iterations by keeping more redundant keypoints, the former loses overlap ratio dramatically after 4 iterations while the latter still keeps the ratio over 90\%.

\textbf{Adaptive sampling with pose uncertainty.} The matching matrix $\mM^{(t)}$ might not be very accurate at the first several iterations when descriptors are not discriminative, impairing the accuracy. To mitigate the problem, we make use of the predicted pose. Note, if most of the predicted matches are correct, the pose is more precise with higher inlier ratio. However, if predicted matches contain many outliers, the pose is probably not accurate and has low inlier ratio. Consequently, we define the uncertainty of the pose on its consistency with matches, as  $r^{(t)}=\frac{|\{(x_i,y_i), s.t. f_{epipolar}(P^{(t)},x_i,y_i) \leq \theta_e\}|}{|\{(x_i,y_i,s_i), s.t. s_i\geq \theta_m\}|}$ ($\theta_e$ is threshold determining inliers). We use $r^{(t)}$ to adjust the sampling threshold $\theta_m$, as $\theta^{(t)}_m=\theta_m * r^{(t)}$, allowing the model to dynamically sample fewer keypoints when both matches and pose are accurate and more when they are not.

\subsection{Iterative process at test time}
\label{sec:test_time}
At test time, after each iteration, we compute matches $\mM^{(t)}$ and estimate the pose from matches with RANSAC~\cite{ransac}. We adopt the relative error between consecutively predicted pose $P^{(t)}$ and $P^{(t-1)}$ as stop criteria to determine if continuing iteration. Specifically, if the maximum angular errors of rotations and translations is less than $\theta_P$, the iteration stops. Benefiting from the embedded geometry information, for about 55\% of the cases in YFCC100m dataset~\cite{yfcc100m}, 6 iterations rather than 9 are enough to find a good pose. The predicted pose is further used for adaptive sampling at test time (Sec.~\ref{sec:dyn_attn}).

Once a good pose is predicted, more matches can be found from the guidance of the pose with function $f_{PM}$. $f_{MP}$ first computes the epipolar distance for all pairs in $\mXt, \mYt$ as $\mMt_{e}$ where $\mMt_{eij}=f_{epipolar}(\mXt_i, \mYt_j)$ and $f_{epipolar} (x,y)=\frac{(y^T\Pt x)^2}{(\Pt x)_1^2 + (\Pt x)_2^2 + (\PtT y)_1^2 + (\PtT y)_2^2}$. Then  $\mMt_{e}$ is binarized with errors smaller than $12px$ as 1 otherwise 0 as $ \bar{\mathcal{M}}^{(t)}_{e}$. Finally, the pose-guided matching matrix is obtained as $\mM^{P} = \mMt \bigodot \bar{\mathcal{M}}^{(t)}_{e}$ ($\bigodot$ is element-wise multiplication). As shown in Fig.~\ref{fig:pose_guided_matching}, matches with high uncertainties caused by similar descriptors at regions with repetitive textures can be effectively mitigated by pose constraints. These matches usually require more number iterations to be found.

%  $f_{PM}$ first computes the Sampson distance~\cite{mvg} for pairs in $\mathcal{X} \times \mathcal{Y}$ with $P^{(t)}$. For each keypoint $x_i \in \mathcal{X}$,  $y_j\in \mathcal{Y}$ with the highest matching score along the epipolar line defined by $x_i$ and $P^{(t)}$ is taken as the candidate.

\section{Experiment Setup}
\label{sec:experiment_setup}

%In this section, we give training details and introduce the datasets and metrics used for evaluation. 
%More implementation details can be found in the \textbf{supplementary material}.

\textbf{Training.} We implement the model in PyTorch~\cite{pytorch} and train it on MegaDepth dataset~\cite{megadepth} for 900k iterations with Adam optimizer~\cite{adam}, batch size of 16, initial learning rate of 0.0001. The learning rate decays with ratio of 0.99996 after 200k iterations and is fixed at 0.00001. We use 1024 SuperPoint~\cite{superpoint} and RootSIFT~\cite{sift,rootsift2012} keypoints for training. As with SuperGlue~\cite{superglue}, our network has $T$ (9) iterative blocks. $\alpha_{m}, \alpha_{p}$, $\alpha_{g}$, $\theta_m$, $\theta_P$, $\theta_e$ are set to 0.6, 0.2, 0.2, 0.2, 1.5 and 0.005, respectively. Note that unlike SuperGlue which is first trained on Oxford-Paris dataset~\cite{oxford2018} and then fine-tuned on Scannet~\cite{scannet} and MegaDepth~\cite{megadepth} datasets for indoor and outdoor models respectively, our models are trained only on the MegaDepth dataset from scratch. %\textbf{We will release source code including the training part for reproducing results}.

\textbf{Datasets and metrics.} We first test our method on YFCC100m~\cite{yfcc100m} and Scannet~\cite{scannet} datasets to evaluate the performance on relative pose estimation. YFCC100m is a large-scale outdoor dataset consisting of images with large illumination and season changes and viewpoint variations. Scannet is an indoor dataset widely used for depth prediction~\cite{probdepth2022} and pose estimation~\cite{superglue,sgmnet}. For relative pose estimation, we report the accurate cumulative error curve (AUC)~\cite{superglue} at thresholds of $5^\circ$, $10^\circ$, and $20^\circ$. The error is the maximum angular errors of rotations and normalized translations. We also report the mean matching score (M.S.) and mean precision (Prec.) of matches.

We additionally test our method on large-scale localization task at Aachen Day-Night (v1.0 and v1.1) datasets~\cite{aachen,aachenv112021}. Aachen v1.0 dataset comprises of 4,328 reference and 922 query (824 day, 98 night) images. Aachen v1.1 extends v1.0 by adding additional 2,369 reference and 93 night query images. We use HLoc~\cite{hloc2019} pipeline for mapping and localization and report the accuracy at error thresholds of $0.25m/2^\circ$, $0.5m/5^\circ$, and $5m/10^\circ$.

\textbf{Baselines.} The baselines comprise of simple matchers such as MNN and NN-RT~\cite{sift}. Also, filtering-based methods including OANet~\cite{oanet}, AdaLAM~\cite{adalam}, CLNet~\cite{clnet} and LMCNet~\cite{lmcnet} are also included. The final part is transformer-based matchers such as SuperGlue~\cite{superglue}, SGMNet~\cite{sgmnet}, and ClusterGNN~\cite{clustergnn}. As it is difficult to reproduce the results of SuperGlue from custom training (official training code is not released), we follow SGMNet and show results of original SuperGlue (SuperGlue*) and the model trained by SGMNet (SuperGlue).
\section{Experiment Results}
\label{sec:experiments}

\setlength{\tabcolsep}{2.5pt}

\begin{table}
	%\footnotesize
	%\small
	\scriptsize
	\centering
		\begin{tabular}{ll|ccc|c|c}
			\toprule
			%\multirow{2}{*}{\textit{Feature}} & \multirow{2}{*}{\textit{Matcher}} & \multicolumn{3}{|c|}{AUC} & \multirow{2}{*}{M.S.} & \multirow{2}{*}{Prec.(\%)}\\
			%& &  $@5^\circ$ & $@10^\circ$ & $20^\circ$ & & \\
			\textit{Feature} & \textit{Matcher} & $@5^\circ$ & $@10^\circ$ & $@20^\circ$ & M.S.(\%) & Prec.(\%)\\
			\midrule
			\multirow{11}{*}{RootSIFT~\cite{sift}}
			
			& NN-RT~\cite{sift} & 26.7 & 43.2 & 59.4 & 4.4 &  56.4 \\
			& AdaLAM (4k)~\cite{adalam}& 27.5 & 44.5  & 60.5 & 6.3 & 84.3   \\
			& OANet~\cite{oanet}~\cite{oanet} & 22.4 & 36.3 & 50.3 & 5.6 & 53.7 \\
			& CLNet~\cite{clnet} & 33.0 & 52.1 & 68.5 & 7.8 & 75.2 \\
			& LMCNet~\cite{lmcnet} & 35.8 & 55.6 & 72.2 & - & 86.9 \\
			
			\cdashline{2-7}
			
			& SuperGlue*~\cite{superglue} & 35.3 & 56.1 & \textbf{73.6} & - & -\\
			
			& SuperGlue~\cite{superglue,sgmnet} & 35.1 & 54.2 & 70.9 & 16.6 & 81.7\\
			& SGMNet\cite{sgmnet} & 34.8 & 54.1 & 70.9 & 17.1 & 86.1 \\
			& ClusterGNN~\cite{clustergnn} &  32.8 & 50.3  & 65.9 & - & -\\
			
			\cdashline{2-7}
			
			%& \textbf{IMP-base} & 36.3 & 56.1 & 72.5 & \underline{17.5} & 84.2 \\
			%& Ours (w/p) & 36.7 & 56.4 & 72.8 & 18.0 & 83.8\\
			&  \textbf{IMP} & \underline{36.7} & \textbf{56.6} & \underline{72.9} & \textbf{18.0} & \underline{87.6} \\
			% &  \textbf{EIMP-base} & 36.4 & 56.1 & 72.5 & 13.8 & \underline{87.3} \\
			%& Ours (ada w/p) & 36.9 & 56.9 & 72.8 & 13.7 & 87.8 \\
			%& Ours (ada w/p) + stop & 36.9 & 56.7 & 72.9 & 13.7 & 79.8 \\
			&  \textbf{EIMP} & \textbf{36.8} & \underline{56.3} & 72.8 & 13.7 & \textbf{89.8} \\
			
			\midrule
			
			\multirow{11}{*}{SuperPoint~\cite{superpoint}} 
			& MNN & 6.5 &  15.4 & 28.5 & 16.2 & 16.2 \\
			& AdaLAM (2k)~\cite{adalam} & 20.8 & 36.5 & 51.9 & 10.9 & 72.0 \\
			& OANet~\cite{oanet} & 19.2 & 34.5 &  50.3 &  9.4 & 62.1\\
			& CLNet~\cite{clnet} & 27.8 & 46.4 & 63.8 & 11.9 & 75.1\\
			& LMCNet~\cite{lmcnet} & 17.4 & 33.2 & 51.1 & - & \textbf{88.9} \\
			
			\cdashline{2-7}
			
			& SuperGlue*~\cite{superglue} & 37.1 & 57.2 & 73.6 & 21.7 & \underline{88.5} \\
			& SuperGlue~\cite{superglue,sgmnet} & 33.2 & 53.5 & 70.8 & 19.7 & 78.7 \\
			& SGMNet~\cite{sgmnet} & 33.0 & 53.0 & 70.0 & \underline{22.3} &  85.1 \\
			& ClusterGNN~\cite{clustergnn} &  35.3 & 56.1 & 73.6 & - & -\\
			
			%&  \textbf{IMP-base} & \underline{38.6} & \underline{58.3} & \underline{74.6} & \textbf{23.6} & 87.2 \\
			%& Ours (w/p) & \textbf{39.1} & \textbf{59.0} & \textbf{75.0} & \textbf{23.7} & 87.7 \\
			\cdashline{2-7}
			
			&  \textbf{IMP} & \textbf{39.4} & \textbf{59.4} & \textbf{75.2} & \textbf{23.0}  & 84.9 \\
			%&  \textbf{EIMP-base} & 36.4 & 56.5 & 73.0 & 18.8 & 88.4 \\
			%& Ours (ada w/p) & 36.9 & 56.6 & 73.3 & 88.7 & 18.4 \\
			%& Ours (ada w/p) + stop & 37.2 & 57.2 & 73.3 & 86.5 & 18.3 \\
			&  \textbf{EIMP} & \underline{37.9} & \underline{57.9} & \underline{74.0} & 19.9 & 88.4 \\ 
			\bottomrule			   
		\end{tabular}
	\caption{\textbf{Results on YFCC100m dataset~\cite{yfcc100m}.} We report the AUC of relative poses at error thresholds of $5^\circ$, $10^\circ$, and $20^\circ$. The mean matching score (M.S.) and mean matching accuracy (Prec.) are also reported. The \textbf{best} and \underline{second best} results are highlighted.}
	\label{tab:yfcc}
\end{table}

In this section, we first show results on relative pose estimation and localization tasks in Sec.~\ref{sec:exp:rel} and Sec.~\ref{sec:exp:localization}. Next, we analyze the computational cost in Sec.~\ref{sec:running_time}. Finally, we conduct a full ablation study to test each component in our framework in Sec.~\ref{sec:ablation}.

\subsection{Relative pose estimation}
\label{sec:exp:rel}

\textbf{Comparison with filtering-based methods.} Table~\ref{tab:yfcc} shows our IMP and efficient IMP (EIMP) give higher accuracy than previous filtering-based approaches such as OANet~\cite{oanet} and LMCNet~\cite{lmcnet} for both RootSIFT~\cite{sift,rootsift2012} and SuperPoint~\cite{superpoint} features. Because these filtering-based methods rely purely on the geometric information to filter potentially wrong correspondences, their performance is influenced heavily by the quality of initial matches. In contrast, we use both the geometric information and descriptors for matching, which are able to give more accurate matches, resulting in more precise poses.

\textbf{Comparison with matchers.} Augmented with spatial information, SuperGlue~\cite{superglue} and its variants~\cite{sgmnet,clustergnn} outperform MNN and NN-RT obviously. Although SGMNet~\cite{sgmnet} and ClusterGNN~\cite{clustergnn} are faster, they give worse accuracy than SuperGlue* due to the loss of information in the message propagation process. With implicitly embedded geometric information, our IMP gives more pose-aware matches and hence outperforms SuperGlue* especially for very precise pose estimation at error threshold of $5^\circ$. Enhanced by geometric information, our EIMP even gives slightly better results than SuperGlue* at error thresholds of $5^\circ$ and $10^\circ$. Our EIMP outperforms previous efficient methods such as SGMNet and ClusterGNN at all error thresholds but has close even higher efficiency.

Table~\ref{tab:scannet} shows results on the Scannet dataset~\cite{scannet}. Our IMP and EIMP also give higher accuracy than filtering-based approaches and other matchers on RootSIFT keypoints. When using SuperPoint, our IMP gives slightly better performance than SuperGlue* and SGMNet, which yield close performance to our EIMP. LMCNet~\cite{lmcnet} reports the best accuracy at error thresholds of $10^\circ$ and $20^\circ$ with SuperPoint because LMCNet uses SuperGlue* to provide initial matches and is further trained on the indoor dataset~\cite{sun3d2013}.

\begin{figure*}[t!]
 	\centering
 	\includegraphics[width=.99\linewidth]{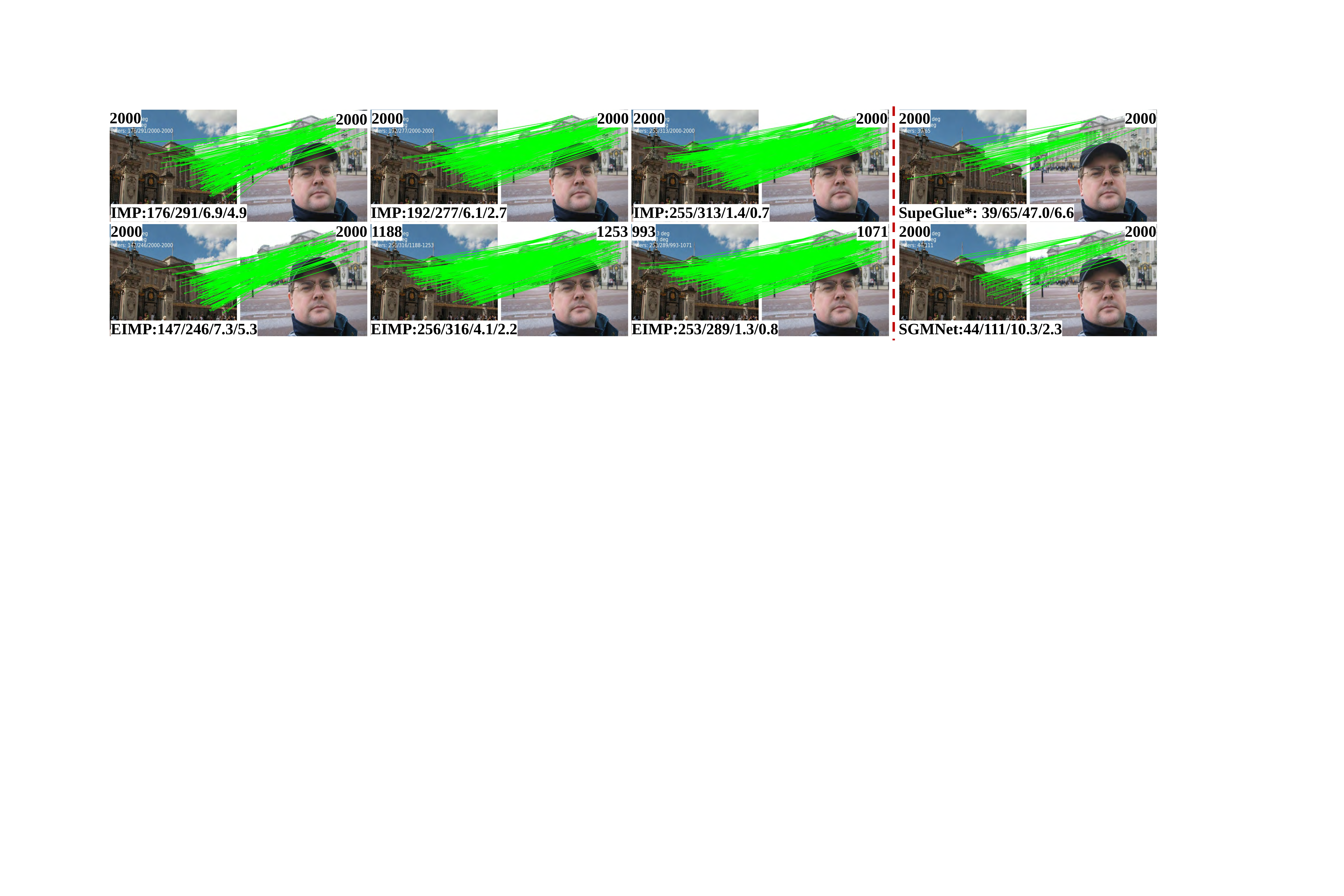}
 	\caption{\textbf{Qualitative comparison.} We visualize the iterative process (left$\rightarrow$right) of IMP (top) and EIMP (bottom). For each pair, we report the number inliers/matches and rotation/translation errors (bottom) and the number of preserved keypoints in two sets (top) at each iteration. \textcolor{green}{Inliers} between two images are visualized. In the iterative process, our model not only finds more inliers but enforces inliers to span wider regions. However, both SuperGlue* and SGMNet predict fewer matches from a small area, resulting higher pose errors.}
 	\label{fig:match_comp}
\end{figure*}

\textbf{Qualitative comparison.} As shown in Fig.~\ref{fig:match_comp}, in the iteration process, IMP and EIMP produce more inliers and more accurate poses progressively. Besides, EIMP dynamically discards useless keypoints after each iteration. We also show the results of SuperGlue*~\cite{superglue} and SGMNet~\cite{sgmnet}. Due to large viewpoint changes, both SuperGlue and SGMNet give much fewer inliers from only a small area and have larger errors compared to our models. In the iterative process, inliers span almost the whole meaningful regions, resulting in more robust pose estimation.
 
\setlength{\tabcolsep}{2.5pt}

\begin{table}
	%\footnotesize
	%\small
	\scriptsize
	\centering
		\begin{tabular}{ll|ccc|c|c}
			\toprule
			%\multirow{2}{*}{\textit{Feature}} & %\multirow{2}{*}{\textit{Matcher}} & \multicolumn{3}{|c|}{AUC} & %\multirow{2}{*}{M.S.} & \multirow{2}{*}{Prec.(\%)}\\
			%& &  $@5^\circ$ & $@10^\circ$ & $20^\circ$ & & \\
			\textit{Feature} & \textit{Matcher} & $@5^\circ$ & $@10^\circ$ & $@20^\circ$ & M.S.(\%) & Prec.(\%)\\
			\toprule
			\multirow{9}{*}{RootSIFT~\cite{sift}} & NN+RT~\cite{sift} & 9.1 & 19.8 & 32.7 & 2.3 & 28.8 \\
			& AdaLAM (4k)~\cite{adalam} & 8.2 & 18.6 & 31.0 & 3.1 & \textbf{47.6} \\
			& OANet~\cite{oanet} & 10.7 &  23.1 & 37.4 & 3.2 & 36.9 \\
			& CLNet~\cite{clnet} & 5.8 & 15.1 & 26.8 & 2.0 & 43.9 \\
			& LMCNet~\cite{lmcnet} & 8.5 & 19.3 & 32.4 & - & \underline{47.0} \\ 
			
			\cdashline{2-7}
			& SuperGlue*~\cite{superglue} & - & - & - & - & -\\
			& SuperGlue~\cite{superglue,sgmnet} & 14.7 & 29.4 & 45.6 & \underline{8.4} & 42.2 \\
			& SGMNet\cite{sgmnet} & 14.4 & 29.9 & 46.0 & \textbf{8.8} & 45.6  \\

			%& \textbf{IMP-base} & 14.6 & 30.7 & \textbf{47.5} & \textbf{9.9} & 42.8 \\
			%& Ours (w/p) & 15.4 & 31.2 & 47.8 & 10.1 & 42.6 \\
			
			\cdashline{2-7}
			
			& \textbf{IMP} & \textbf{15.6} & \textbf{30.9} & \textbf{47.4} & 5.8 & 42.0 \\
			%& \textbf{EIMP} & 14.5 & \textbf{30.9} & 47.3 & 7.7 & 45.8 \\
			%& Ours (ada w/p) & 14.7 & 30.4 & 47.3 & 7.5 & 46.2 \\
			%& Ours (ada w/p) + early & 15.0 & 30.5 & 46.8 & 5.3 & 42.3 \\
			& \textbf{EIMP} & \underline{15.3} & \underline{30.8} & \underline{46.6} & 7.7 & 45.7\\ 
			\midrule
			
			\multirow{10}{*}{SuperPoint~\cite{superpoint}} 
			%& MNN & 5.0 & 12.8 & 24.3 & 17.1 & 17.1\\
			%& SuperGlue*~\cite{superglue} & 15.4 & 31.9 & 49.3 & 16.8 & 47.9 \\
			%& SuperGlue~\cite{superglue} &  12.4 & 26.6  & 42.4 & 15.0 & 45.9  \\
			%& SGMNet~\cite{sgmnet} & 16.3 & 31.8 & 48.7 & 17.0 & 48.2 \\
			% results from SGMNet
			& MNN & 9.4 & 21.6 & 36.4 & 13.3 & 30.2\\
			& AdaLAM (2k)~\cite{adalam} &  6.7 & 15.8 & 27.4 & 13.2 & 44.2  \\
			& OANet~\cite{oanet} &  10.0 & 25.1 &  38.0 &  10.6 & 44.6 \\
			& CLNet~\cite{clnet} & 4.1 & 11.0 & 21.6 & 8.6 & 44.2\\
			& LMCNet~\cite{lmcnet} & 14.6 & \textbf{33.6} & \textbf{53.6} & - & 36.8\\
			
			\cdashline{2-7}
			
			& SuperGlue*~\cite{superglue} & 16.2 & 32.6 & 49.3 & \underline{16.8} & \underline{47.9} \\
			& SuperGlue~\cite{superglue,sgmnet} &  12.0 & 26.3 & 42.4 & 15.0 & 45.9  \\
			& SGMNet~\cite{sgmnet} & \underline{16.4} & 32.1 & 48.7 & \textbf{17.0} & \textbf{48.1} \\
			
			%& \textbf{IMP-base} & \textbf{16.6} & 32.4 & 48.7 & \textbf{17.0} & 45.9 \\
			%& Ours (w/p) & 16.5 & 32.5 & 49.1 & 17.1 & 46.2 \\
			\cdashline{2-7}
			
			& \textbf{IMP} & \textbf{16.6} & \underline{33.1} & \underline{49.4} & 15.8 & 42.0 \\
			%&  \textbf{EIMP-base} & 15.7 & 31.8 & 47.8 & 15.3 & 46.9 \\
			%& Ours (ada + w/p) & 14.7 & 30.4 & 47.3 & 7.5 & 46.2 \\
			%& Ours (ada + w/p) + stop & 15.0 & 30.5 & 46.8 & 5.3 & 42.3 \\
			& \textbf{EIMP} & 15.9 & 32.4 & 48.9 & 15.9 & 46.2\\
			\bottomrule			   
		\end{tabular}
	\caption{\textbf{Results on Scannet dataset~\cite{scannet}.} We show the AUC of relative poses at error thresholds of $5^\circ$, $10^\circ$, and $20^\circ$. The mean matching score (M.S.) and mean matching accuracy (Prec.) are also reported. The \textbf{best} and \underline{second best} results are highlighted.}
	\label{tab:scannet}
\end{table}

\subsection{Visual localization}
\label{sec:exp:localization} 
As most filtering methods don't provide results on visual localization task, we only show results of OANet~\cite{oanet} and AdaLAM~\cite{adalam}. OANet and AdaLAM give close accuracy to MNN at error thresholds of $0.25m, 2^\circ$ and $0.5m, 5^\circ$, but much better performance at threshold of $5m, 10^\circ$, because test images with larger viewpoint changes are more sensitive to outliers which can be partially filtered by OANet and AdaLAM. Our IMP obtains similar performance to SuperGlue*~\cite{superglue} and outperforms SGMNet and ClusterGNN especially for night images. Note that our models are only trained on the MegaDepth dataset~\cite{megadepth} while SuperGlue* is additionally pretrained on Oxford and Pairs dataset~\cite{oxford2018}.

EIMP slightly outperforms IMP and SuperGlue*. That is because in the long-term large-scale localization task, query and reference images usually have larger viewpoint and illumination changes with more keypoints without true matches. EIMP effectively discards these keypoints, guaranteeing the quality of matches and thus improving the localization accuracy. %Similar results can also be observed in Fig.~\ref{fig:match_comp} where EIMP also boots the relative pose estimation at the first and second iterations.

\setlength{\tabcolsep}{3.5pt}

\begin{table}
	%\footnotesize
	%\small
	\scriptsize
	\centering
		\begin{tabular}{ll|cc}
			\toprule
			\textit{Feature} & \textit{Matcher} & Day & Night \\
			& & \multicolumn{2}{c}{$(0.25m,2^\circ) / (0.5m,5^\circ) / (5m,10^\circ)$}\\
			\midrule
			
			\multirow{9}{*}{SuperPoint~\cite{superpoint}} & MNN & 85.4 / 93.3 / 97.2 & 75.5 / 86.7 / 92.9 \\
			& OANet~\cite{oanet} & - & 77.6 / 86.7 / 98.0\\
			& AdaLAM~\cite{adalam} & - & 78.6 / 86.7 / 98.0 \\
			%& SuperGlue*~\cite{superglue} & 79.6 & 90.8 & 100.0 \\
			\cdashline{2-4}
			
			& ELAM~\cite{ela2022} & - & 78.6 / 87.8 / 96.9 \\ 
			& SuperGlue~\cite{superglue,sgmnet} & - & 76.5 / 88.8 / 99.0 \\
			& SuperGlue*~\cite{superglue} & \underline{89.6} / 95.4 / 98.8 & \textbf{86.7} / \underline{93.9} / \textbf{100.0}\\
			& SGMNet~\cite{sgmnet} & 86.8 / 94.2 / 97.7 & 83.7 / 91.8 / \underline{99.0} \\
			& ClusterGNN~\cite{clustergnn} &89.4 / \underline{95.5} / 98.5 & 81.6 / \underline{93.9} / \textbf{100.0}\\
			
			\cdashline{2-4}
			
			%& Ours & 89.4 / 95.4 / 99.0 & 86.7 / 93.9 / 100.0  \\
			& \textbf{IMP} & 89.1 / 95.4 / \underline{99.0} & \textbf{86.7} / \textbf{94.9} / \textbf{100.0}\\
			%& Ours (ada) & 90.2 / 96.2 / 99.0 & 84.7 / 93.9 / 100.0 \\
			& \textbf{EIMP} & \textbf{90.0} / \textbf{96.5} / \textbf{99.2} & \underline{84.7} / \textbf{94.9} / \textbf{100.0} \\
			
			\midrule
			\multirow{5}{*}{SuperPoint~\cite{superpoint}}
			& MNN & 87.9 / 93.6 / 96.8 & 70.2 / 84.8 / 93.7 \\
			& AdaLAM~\cite{adalam} & - & 73.3 / 86.9 / 97.9 \\
			& SuperGlue*~\cite{superglue} & \underline{89.8} / \textbf{96.6} / \textbf{99.4} & 75.9 / 90.1 / \textbf{100.0} \\
			& SGMNet~\cite{sgmnet} & 88.7 / 96.2 / 98.9 & 75.9 / 89.0 / 99.0 \\
			\cdashline{2-4}
			
			& \textbf{IMP} & 89.1 / 95.4 / \underline{99.0} & 75.9 / \textbf{92.7} / \underline{99.5} \\
			& \textbf{EIMP} & \textbf{90.0} / \underline{96.5} / \underline{99.0} & \textbf{77.0} / \underline{91.6} / \underline{99.5}\\ 
			\bottomrule			   
		\end{tabular}
	\caption{\textbf{Results on Aachen v1.0 (top) and v1.1 (bottom) dataset~\cite{aachen,aachenv112021}}. The \textbf{best} and \underline{second best} results are highlighted.}
	\label{tab:aachen}
\end{table}

\subsection{Running time}
\label{sec:running_time}

\begin{figure}[t]
	\centering
	\begin{subfigure}{.22\textwidth}
		\centering
		\includegraphics[width=1.\linewidth]{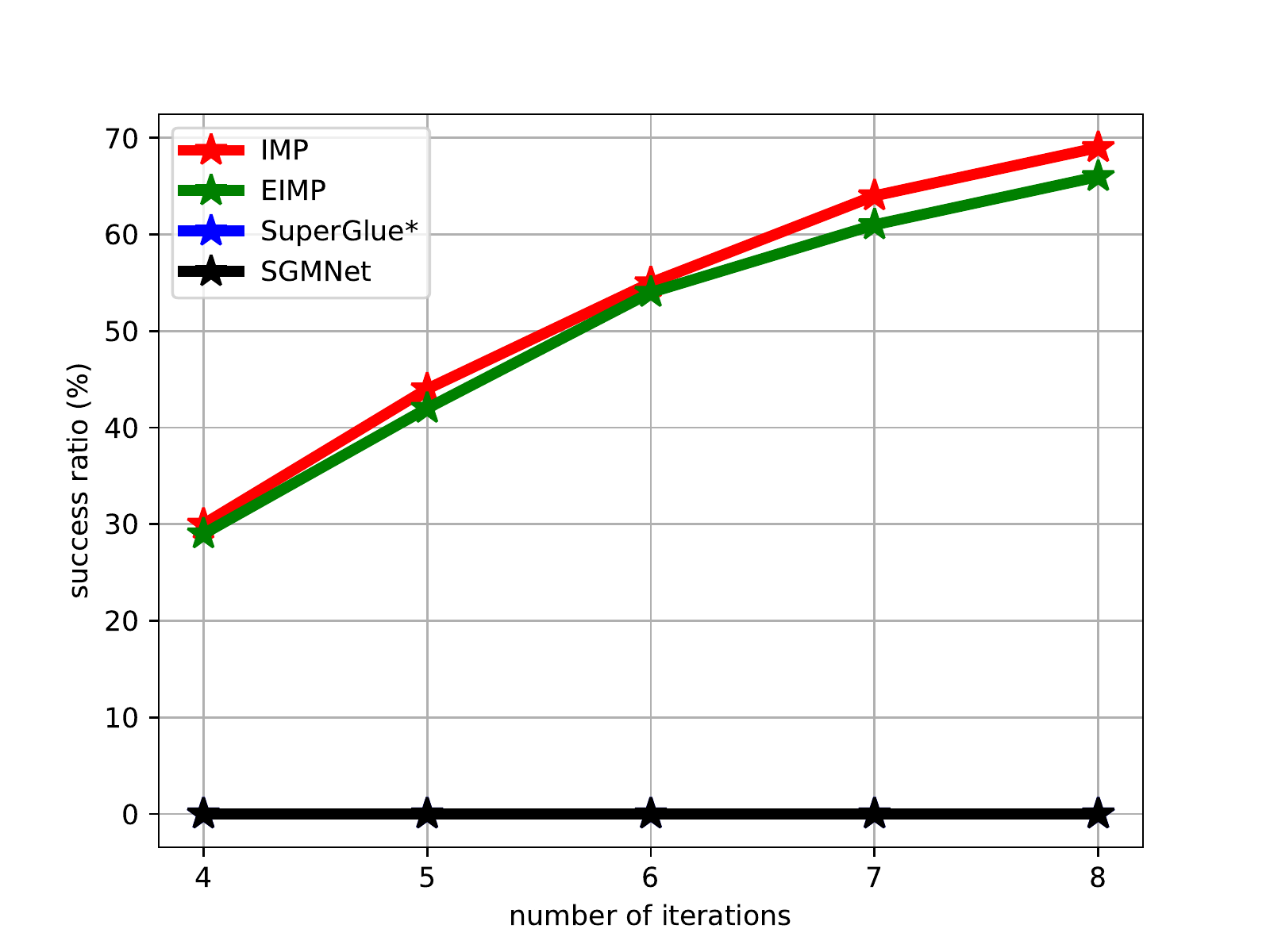}
		\caption{}
		\label{fig:num_iteration}
	\end{subfigure}\quad
	\begin{subfigure}{.22\textwidth}
		\centering
		\includegraphics[width=1.\linewidth]{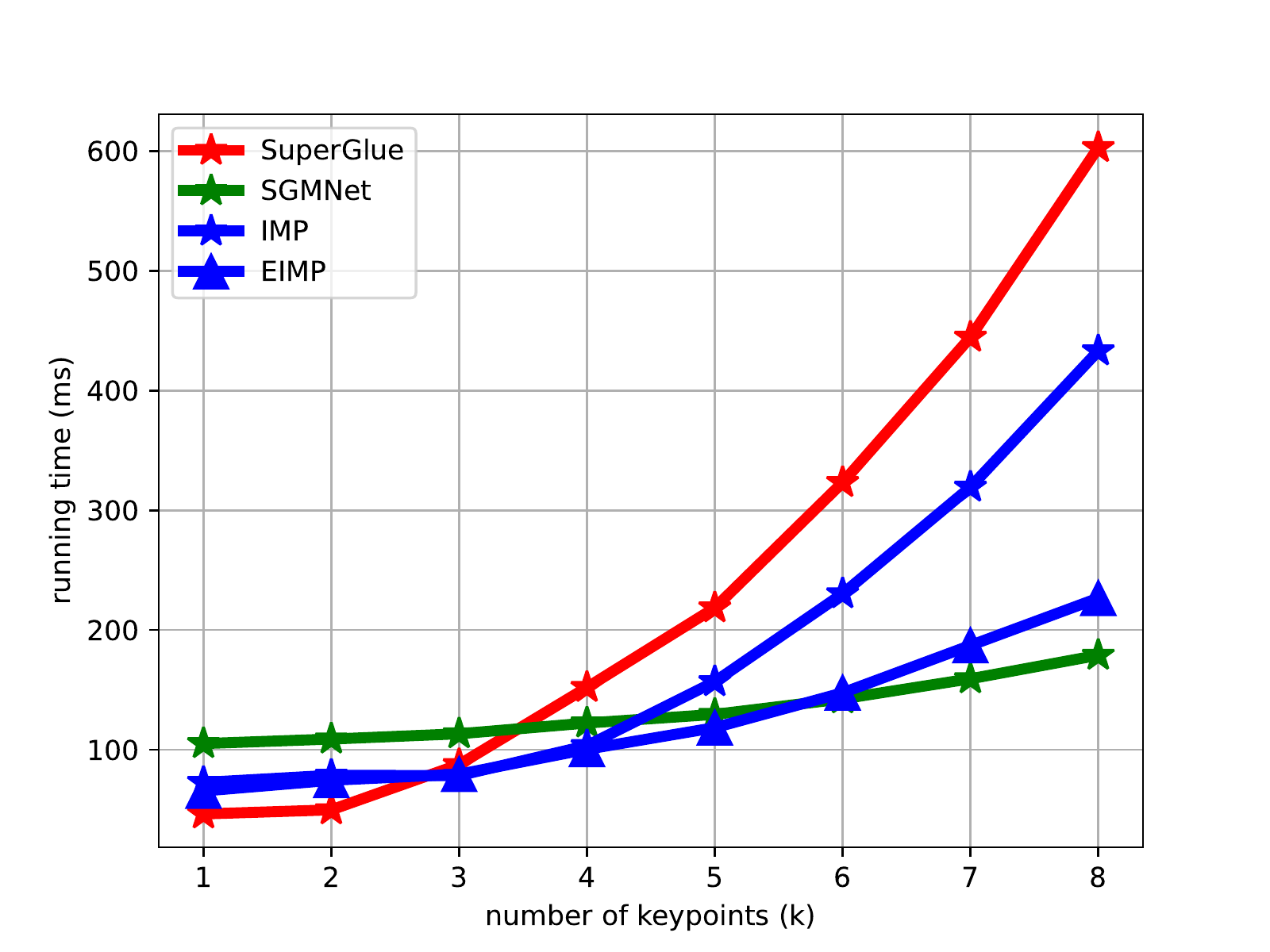}
		\caption{}
		\label{fig:total_time}
	\end{subfigure}
	\vspace*{-10pt}
	\caption{\textbf{Running time analysis.} We show (a) the number of iterations required for each pair YFCC100m dataset~\cite{yfcc100m} and (b) the mean running time of 1k pairs on RTX 3090.}
	\label{fig:running_time}
	\vspace*{-10pt}
\end{figure}
 
As the source code of ClusterGNN~\cite{clustergnn} is not released, we mainly compare our method with SuperGlue*~\cite{superglue} and SGMNet~\cite{sgmnet}. The total time for each method is the summary of the time used at each iteration. IMP reduces the time by reducing the number of iterations and EIMP further decreases the time of each iteration.
 
Fig.~\ref{fig:num_iteration} shows the number of iterations for relative pose estimation on YFCC100m dataset~\cite{yfcc100m}. SuperGlue* and SGMNet adopt a fixed number of layers and only give matches at the last layer, so they don't report any results before the last iteration. In contrast, IMP needs only 4 iterations for 30\% cases and 5 iterations for 55\% cases to find a good pose (relative pose error less than 1.5), avoiding extra computation. Although EIMP loses some matches by discarding useless keypoints, it obtains close results to IMP. Fig.~\ref{fig:total_time} shows the running time of SuperGlue*, SGMNet, IMP and EIMP. When using 1k and 2k keypoints, IMP and EIMP run slower than SuperGlue* because of additional in shared attention. However, when using more than 3k keypoints, IMP runs faster because it uses fewer number of iterations (10) for sinkhorm optimization~\cite{sinkhorn1967,sinkhorn2013}. Benefiting from adaptive pooling strategy, EIMP works even faster than SGMNet when using less than 6k keypoints but gives better accuracy, as demonstrated in Table~\ref{tab:yfcc}.

\subsection{Ablation study}
\label{sec:ablation}
We conduct a full ablation study for relative pose estimation on YFCC100m dataset~\cite{yfcc100m} with SuperPoint~\cite{superpoint} keypoints to verify all components in our model including the shared attention (S), pose-consistency loss (C), pose-aware iteration (P), adaptive pooling (A), and adaptive pooling with pose uncertainty (U).

Table~\ref{tab:ablation_yfcc} shows that shared attention (S) improves both the pose and matching accuracy by propagating more message to each keypoint. By infusing the geometric information in the iterative process, pose-consistency loss (C) and pose-aware iteration (P) also enhance the performance of IMP and EIMP. Comparisons between IMP-SC and EIMP-SCA indicate that our adaptive pooling loses 5\% matches (M.S. 23.7 vs. 18.4) and gains 1\% inliers (Prec. 87.7 vs. 88.7). The pose uncertainty (U) further increases the number of matches (M.S. 18.3 vs. 19.9) and inliers (Prec. 86.5 vs. 88.4) and enhances the pose accuracy about 0.7\% by dynamically adjusting the sampling ratio to avoid over-pruning at each iteration. EIMP-S, which is a ratio-based sampling, gives promising pose accuracy but loses over 10\% matches (M.S. 7.8 vs. 18.4) than adaptive pooling, EIMP-SA, which effectively mitigates this problem.

\setlength{\tabcolsep}{2.5pt}

\begin{table}
	%\footnotesize
	%\small
	\scriptsize
	\centering
		\begin{tabular}{l|ccccc|ccc|c|c}
			\toprule
			Model & S & C & P & A & U & $@5^\circ$ & $@10^\circ$ & $20^\circ$ & M.S.(\%) & Prec.(\%)  \\
			\midrule
			IMP & \xmark & \xmark & \xmark & \xmark & \xmark & 36.9 & 57.1 & 73.7 & \textbf{23.7} & 85.2 \\
			IMP-S (base) & \cmark & \xmark & \xmark & \xmark & \xmark & 38.6 & 58.3 & 74.6 & \underline{23.6} & \underline{87.2} \\
			IMP-SC & \cmark & \cmark & \xmark & \xmark & \xmark  & \underline{39.1} & \underline{59.0} & \underline{75.0} & \textbf{23.7} & \textbf{87.7} \\
			IMP-SCP (full) & \cmark & \cmark & \cmark & \xmark & \xmark & \textbf{39.4} & \textbf{59.4} & \textbf{75.2} & 23.0  & 84.9 \\
			\midrule
			EIMP-S & \cmark & \xmark & \xmark & \xmark & \xmark & 35.4 & 55.5 & 72.1 & 7.8 & \textbf{91.2} \\
			EIMP-SA (base) & \cmark & \xmark & \xmark & \cmark & \xmark & 36.4 & 56.5 & 73.0 & \underline{18.4} & \underline{88.8} \\
			EIMP-SCA & \cmark & \cmark & \xmark & \cmark & \xmark & 36.9 & 56.6 & \underline{73.3}  & \underline{18.4} & 88.7 \\
			EIMP-SCPA & \cmark & \cmark & \cmark & \cmark & \xmark & \underline{37.2} & \underline{57.2} & \underline{73.3} & 18.3 & 86.5  \\
			EIMP-SCPAU (full) & \cmark & \cmark & \cmark & \cmark & \cmark & \textbf{37.9} & \textbf{57.9} & \textbf{74.0} & \textbf{19.9} & 88.4 \\ 
			\bottomrule			   
		\end{tabular}
	\caption{\textbf{Ablation study.} We test the efficacy of shared attention (S), pose-consistency loss (C), adaptive pooling (A), pose-aware iteration (P), and adaptive sampling with pose uncertainty (U) on YFCC100m dataset~\cite{yfcc100m} with SuperPoint~\cite{superpoint} features. The \textbf{best} and \underline{second best} results are highlighted.}
	\label{tab:ablation_yfcc}
\end{table}
%Table~\ref{tab:ablation_aachen} shows the efficacy of geometric loss to the visual localization task. With geometric loss, both the full and efficient models give higher accuracy, while the improvement is relatively small compared to the relative pose estimation in Table~\ref{tab:ablation_yfcc}. 
%\section{Limitations}
%\label{sec:limitation}

\section{Conclusions}
\label{sec:conclusion}

In this paper, we propose the iterative matching and pose estimation framework, allowing the two tasks to boost each other and thus improving the accuracy and efficiency. Particularly, we embed the geometric information into the matching module, enforcing the model to predict matches which are not only accurate but also able to give a good pose. Moreover, in each iteration, we utilize the predicted matches, relative pose, and attention scores to remove keypoints without potential true matches adaptively at each iteration, improving the efficiency and preserving the accuracy. Experiments demonstrate that our method achieves better performance than previous approaches on relative pose estimation and large-scale localization tasks and has high efficiency as well.

\section*{Acknowledgment}
This project is supported by Toyota Motor Europe.

\appendix
\section*{\centering Supplementary Material}

In the supplementary material, we provide additional implementation details, qualitative results and analysis in Sec.~\ref{sec:imp} and Sec.~\ref{sec:result}, respectively.

\section{Implementation}
\label{sec:imp}

\subsection{Training} 
We train our model on the MegaDepth dataset~\cite{megadepth} from scratch. Following SuperGlue~\cite{superglue}, we use 153 scenes with 130k images in total for training and 36 for validation. For each category, we first detect 4096 keypoints for all images and then build correspondences for image pairs with overlap ratio from 0.3-1.0. Matches with re-projection errors less than 5px are deemed as inliers, resulting in different number of inliers for different pairs. In the training process, for each epoch, we randomly choose 80 pairs for each scene. For each pair, we randomly choose 1024 keypoints with inliers ranging from 32 to 512 between two images, respectively. We observe that samples with high inlier ratios boost the convergence and those with low inlier ratios enhance the ability of models for finding matches for tough cases at test time.

Our matching loss is identical to the assignment loss of SGMNet~\cite{sgmnet}, which is more stable than the original version in SuperGlue~\cite{superglue}, as analyzed in the SGMNet paper. These modifications enable us to train the model on the MegaDepth dataset~\cite{megadepth} from scratch without requiring any pretraining.

\subsection{Architecture} 
\begin{figure}[t]
	\centering
	\begin{subfigure}{.25\textwidth}
		\centering
		\includegraphics[width=.99\linewidth]{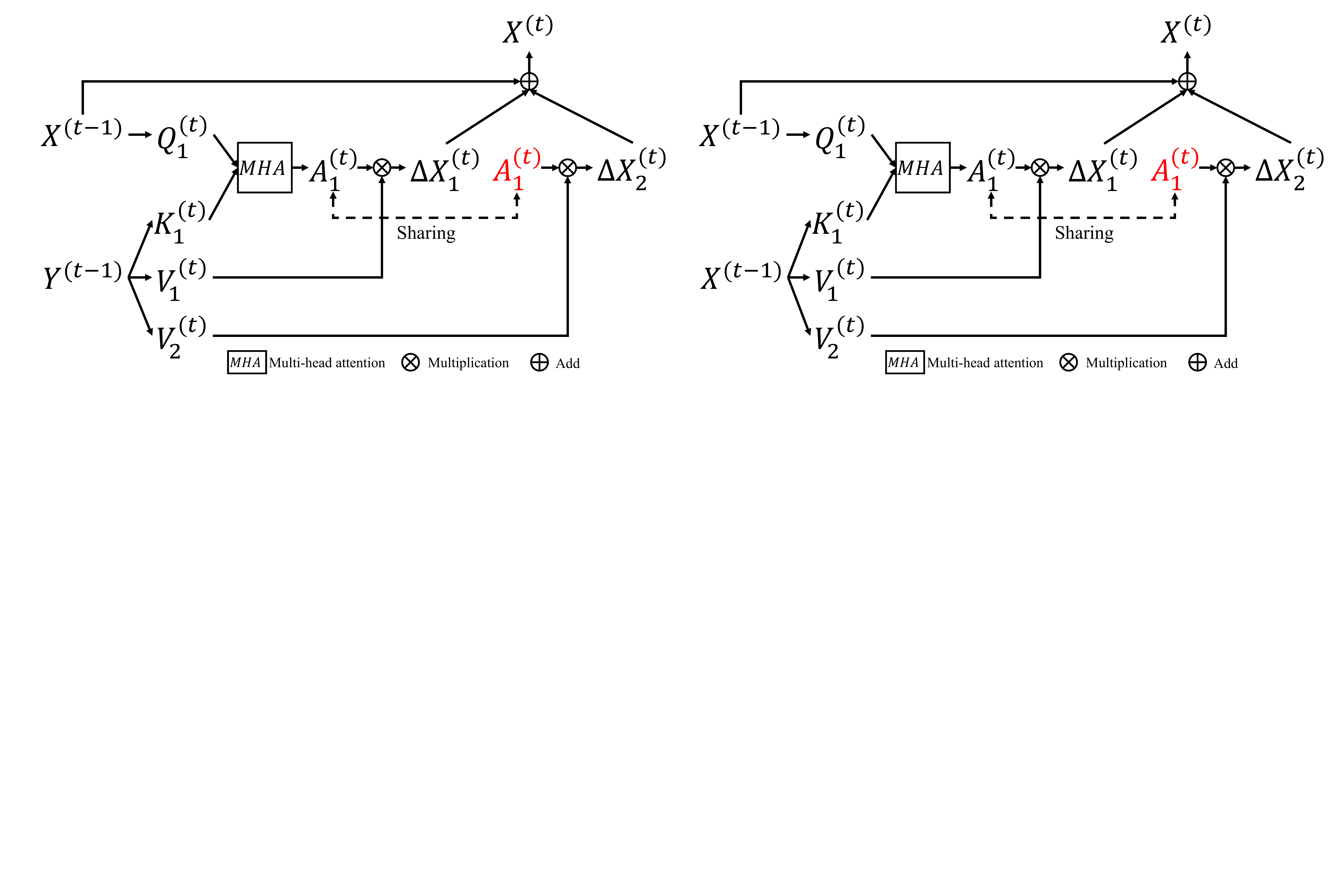}
		\caption{Self attention }
		%\label{fig:pipeline}
	\end{subfigure}%
	\begin{subfigure}{.25\textwidth}
		\centering
		\includegraphics[width=.99\linewidth]{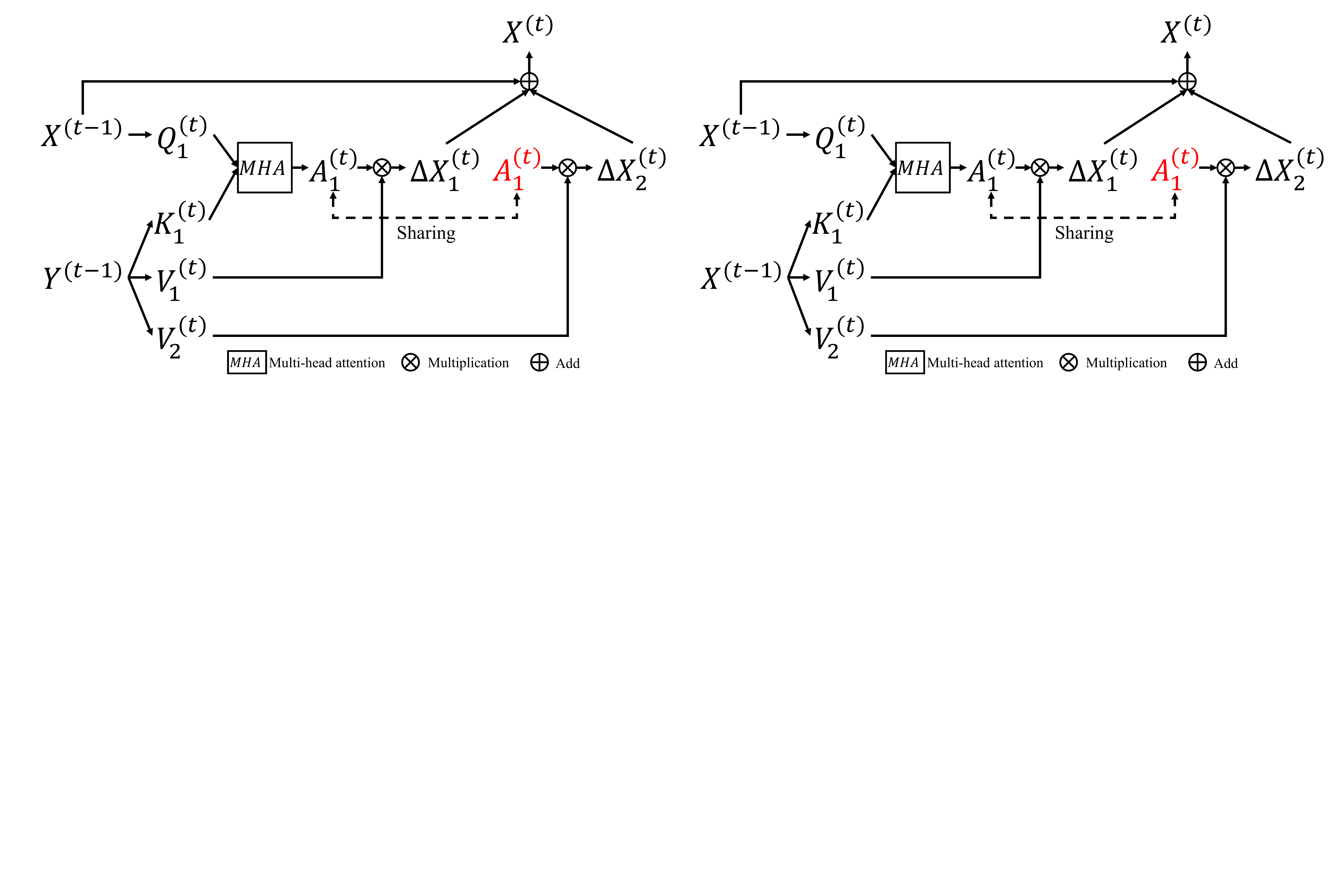}
		\caption{Cross attention}
		%\label{fig:block}
	\end{subfigure}%
	\caption{\textbf{Architecture.} Detailed architecture of self (a) and cross (b) attention with sharing attention matrix.}
	\label{fig:architecture}
\end{figure}

As~\cite{superglue,sgmnet,clustergnn}, we use self and cross attention to gather global information for each keypoint in two sets. Considering the similarities of attention matrices in two consecutive iterations, we adopt the shared attention mechanism~\cite{shareattention} to speed up the message propagation process at low cost. A detailed architecture of self and cross attention is shown in Fig.~\ref{fig:architecture}. We adopt the identical position encoder as SuperGlue~\cite{superglue}.

\subsection{Inference}
At test time, for relative pose estimation on YFCC100m~\cite{yfcc100m} and Scannet~\cite{scannet} datasets, we use the identical testing pairs and number of keypoints for evaluation as SuperGlue~\cite{superglue} and SGMNet~\cite{sgmnet}. As for the metrics, we report the \textbf{exact} cumulative error curve (AUC) rather than the \textbf{approximate} one, because the former measures the error based on groundtruth poses while the latter does it based on predicted poses. 

When evaluating our model on Aachen Day-Night v1.0 and v1.1 datasets~\cite{aachen,aachenv112021}, we adopt HLoc~\cite{hloc2019} pipeline for mapping and localization, as previous methods~\cite{superglue,sgmnet,clustergnn}. We leverage NetVLAD~\cite{netvald2016} to provide 50 reference images for each query image in the localization process.

In our adaptive sampling process, in order to avoid losing too many potential inliers or informative keypoints, we set the minimum number of preserved keypoints for each image to 256. Once the number of keypoints is smaller or equal to 256, we don't perform any sampling. This strategy is also applied to the ratio-based sampling (R50).

\section{Results}
\label{sec:result}

In this section, we first provide more ablation studies in Sec.~\ref{sec:rec:ablation}. Then, we show and discuss more qualitative results of SuperGlue* (official SuperGlue)~\cite{superglue}, SGMNet~\cite{sgmnet} and our IMP and EIMP on Scannet~\cite{scannet}, YFCC100m~\cite{yfcc100m}, and Aachen Day-Night datasets~\cite{aachen,aachenv112021} in Sec.~\ref{sec:rec:scannet}, Sec.~\ref{sec:rec:yfcc}, and Sec.~\ref{sec:rec:aachen}, respectively.

\subsection{Ablation study}
\label{sec:rec:ablation}

\begin{figure}[t]
	\centering
	\begin{subfigure}{.16\textwidth}
		\centering
		\includegraphics[width=.95\linewidth]{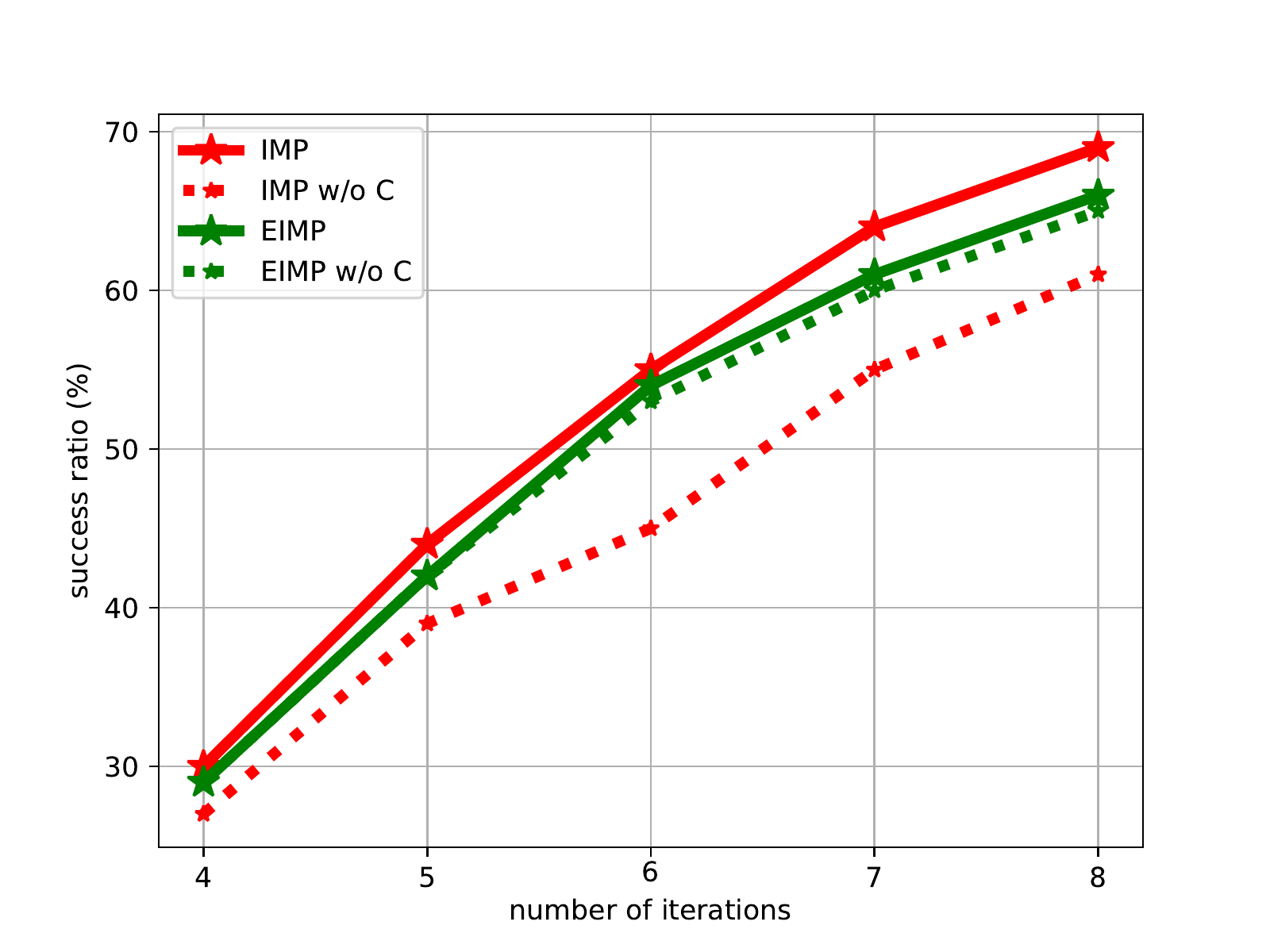}
		\caption{Number of iterations}
		\label{fig:abl_iteration}
	\end{subfigure}%
	\begin{subfigure}{.16\textwidth}
		\centering
		\includegraphics[width=.95\linewidth]{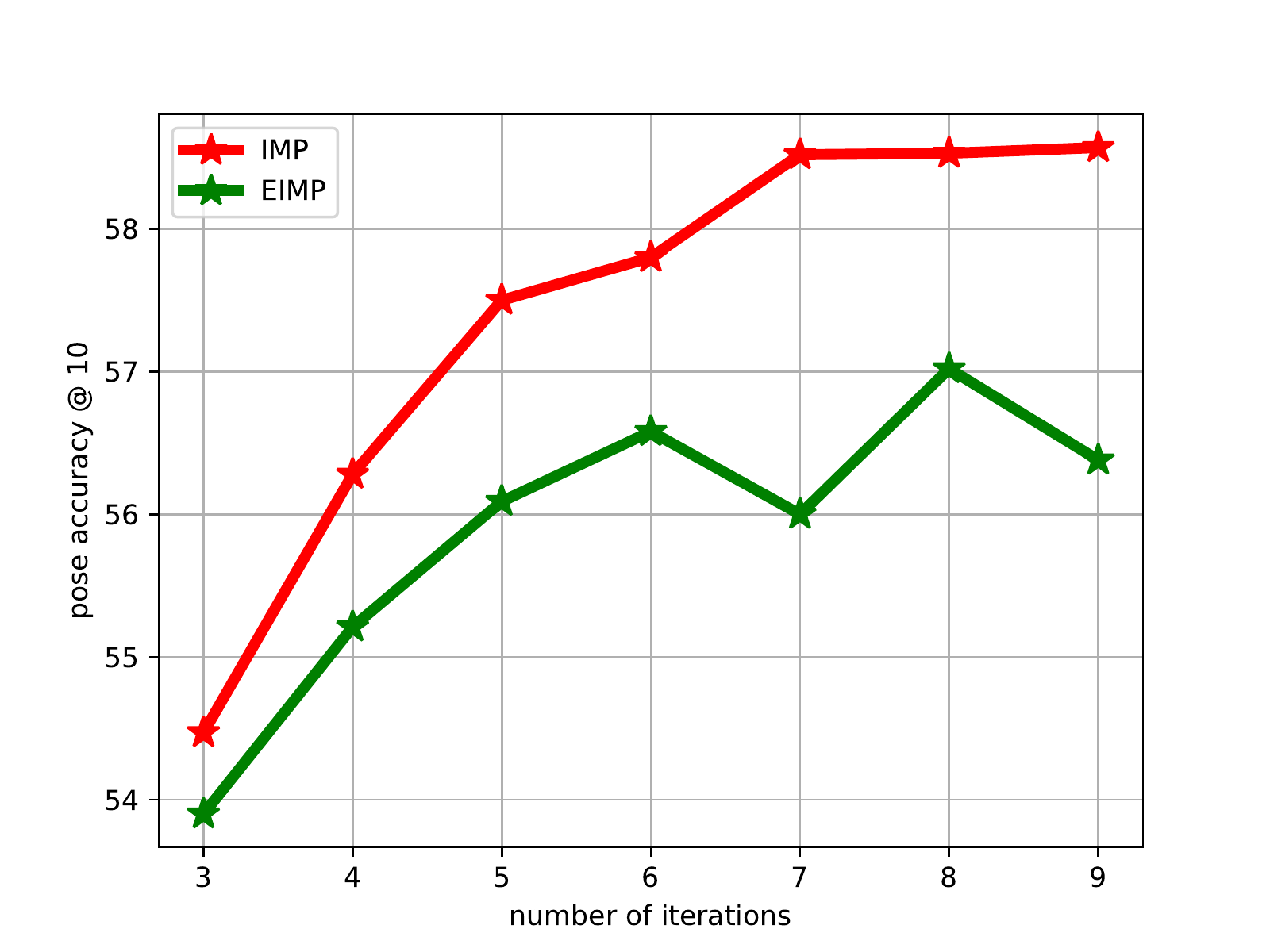}
		\caption{Pose accuracy}
		\label{fig:pose10_iteration}
	\end{subfigure}%
	\begin{subfigure}{.16\textwidth}
		\centering
		\includegraphics[width=.95\linewidth]{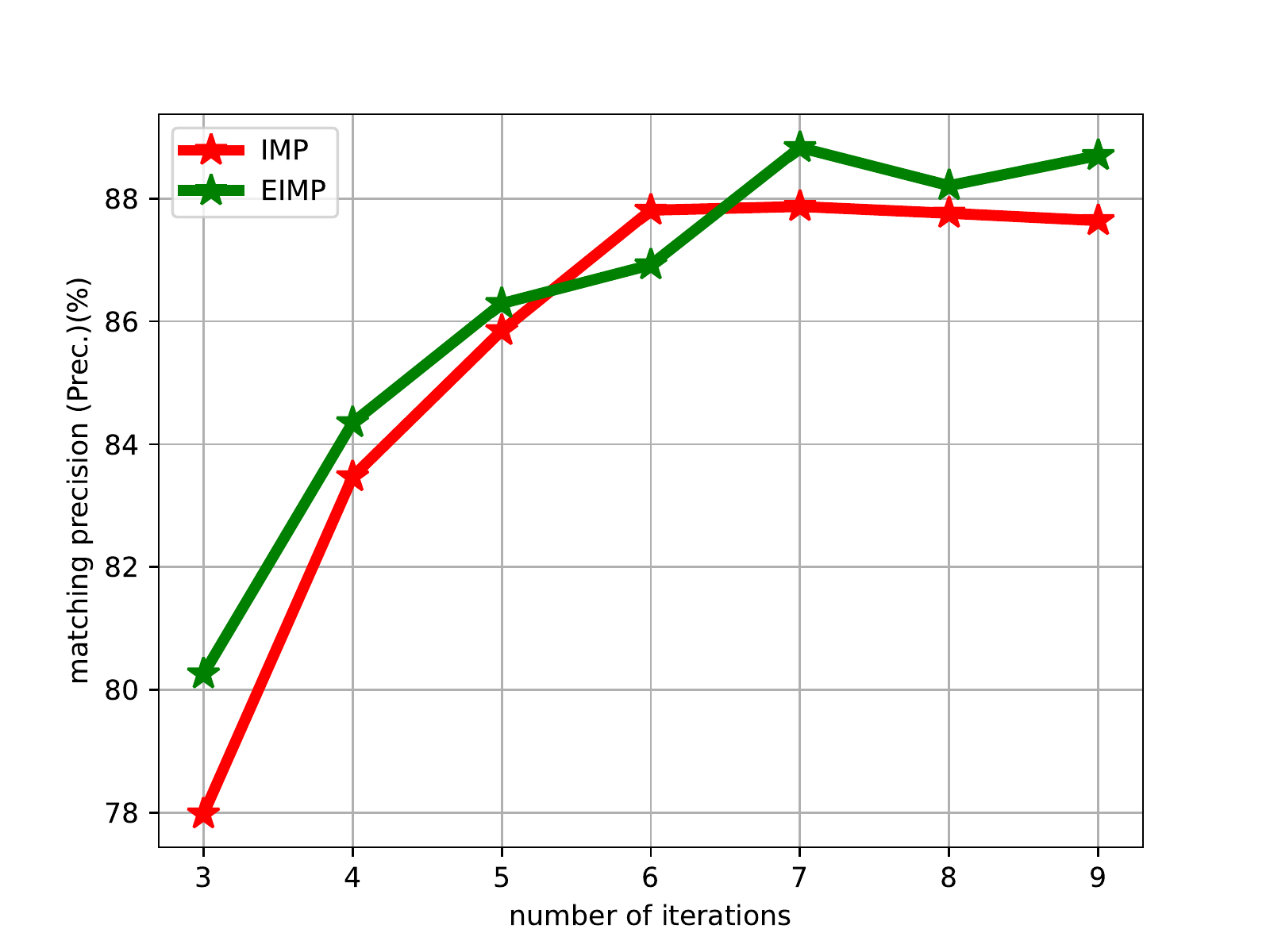}
		\caption{Precision of matches}
		\label{fig:prec_iteration}
	\end{subfigure}%
	\caption{We report the number of iterations for IMP and EIMP with and without pose-consistency (w/o C) loss (a), relative pose accuracy @ $10^\circ$ (b), and precision of matches (c) of IMP and EIMP using different number of iterations on YFCC100m dataset~\cite{yfcc100m}.}
	\label{fig:pose_prec_iteration}
\end{figure}

\textbf{Pose-consistency loss (C).} Pose-consistency loss forces our model to predict matches which are not only correct but also able to give a good pose by implicitly embedding geometric information into the matching module. Fig.~\ref{fig:abl_iteration} shows the influence of pose-consistency loss to the number of iterations required to find a good pose. With pose-consistency loss, as expected, when using the same number of iterations, IMP gives higher success ratios than IMP without this loss (IMP w/o C) because pose-consistency loss allows pose-aware matches pump out first to join the pose estimation, reducing the number of iterations. EIMP adopts the sampling strategy which filters some unreliable matches, therefore the pose-consistency has little influence on the success ratios of EIMP.

\textbf{Number of iterations.} A our model is able to predict matches at each iteration. We test its performance on relative pose estimation by progressively increasing the number of iterations from 3 to 9. As the number of iterations increases, keypoints become more discriminative with more geometric information embedded, so both IMP and EIMP achieve more precise poses, as shown in Fig.~\ref{fig:pose10_iteration}. Additionally, their matching precision (Prec.) also gets higher gradually, indicating that they find more inliers, as shown in Fig.~\ref{fig:prec_iteration}. Fig.~\ref{fig:pose10_iteration} and Fig.~\ref{fig:prec_iteration} show that both IMP and EIMP report almost the best performance when using 7 or 8 iterations, which means IMP and EIMP could even run faster by reducing the maximum number iterations from 9 to 7 or with marginal performance loss. Note that both IMP and EIMP are trained to predict matches at each iteration, so we don't need to retrain or fine-tune IMP and EIMP to achieve this.

%\textbf{Relative pose estimation in localization task.} The localization task needs only the predicted matches to conduct PnP~\cite{epnp2009} + RANSAC~\cite{ransac}, so in the paper we only provide matches and don't use the relative pose between the query and reference images. Here, we also conduct an ablation study to verify if performing relative pose estimation influences the final localization accuracy. Results in Table~\ref{tab:ablation_aachen} demonstrate that if removing relative pose estimation between the query and reference images, our EIMP (w/o P) doesn't lose accuracy mainly because the geometric information is already explicitly embed into the model via pose-consistency loss in the training process. The slightly loss of accuracy of introducing the relative pose estimation could be caused by the losing of inliers as relative pose estimation usually filters a few potential inliers. According to our analysis in the main paper, the iterative relative pose estimation is more important to the essential matrix estimation.

\subsection{Qualitative results on Scannet}
\label{sec:rec:scannet}

\begin{figure*}[t]
	\centering
	\includegraphics[width=.95\linewidth]{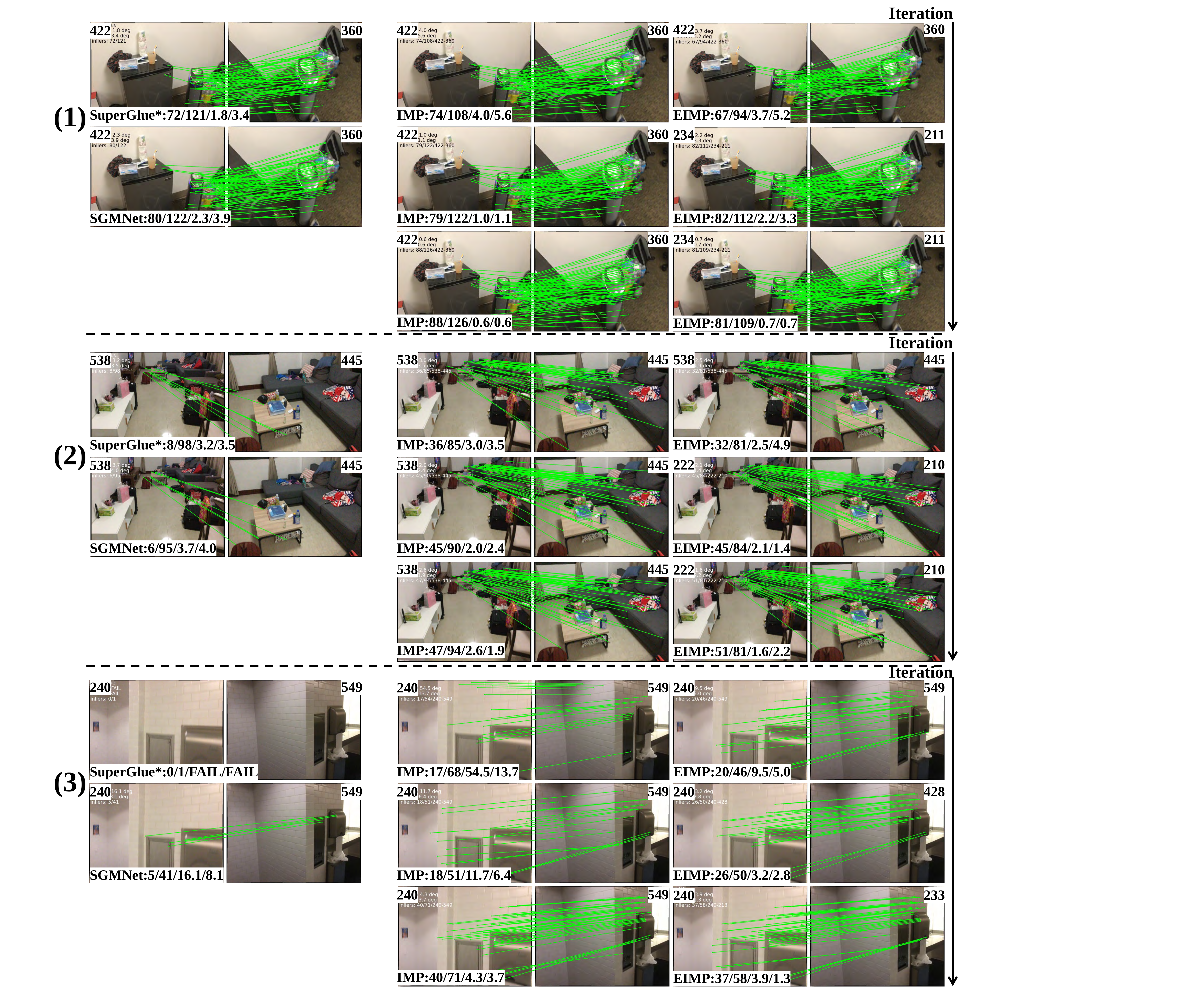}
	\caption{\textbf{Qualitative results on Scannet dataset~\cite{scannet}.} At the left-bottom of each image pair, we report the number of inliers/matches and rotation/translation errors of SuperGlue*~\cite{superglue} (official SuperGlue), SGMNet~\cite{sgmnet}, our IMP and EIMP. Besides, the number of keypoints in each image are shown at the top of each pair. For simple case (1), all methods give similar numbers of inliers, but IMP and EIMP obtain smaller rotation and translation errors than SuperGlue* and SGMNet because of the embedded geometric information. In the iteration process, rather than finding inliers from a cluster, both IMP and EIMP expand the areas with inliers, which allows our models to find more potential inliers and make the pose estimation more stable. When testing images become more difficult ((2), (3)), the behavior of our models in expanding inliers over the whole meaningful regions of images can be observed more clearly. Especially for (3) where all keypoints are extracted from regions with repetitive textures, matching methods based on descriptors can hardly discriminate these keypoints, so SuperGlue* and SGMNet fail to give enough inliers. At this time, geometric constraints play an indispensable role at finding correct matches, which explains the success of IMP and EIMP. }
	\label{fig:vis_scannet}
\end{figure*}

In Fig.~\ref{fig:vis_scannet}, we visualize predicted matches and relative poses of SuperGlue*~\cite{superglue}, SGMNet~\cite{sgmnet}, our IMP and EIMP. We observe that for simple cases (Fig.~\ref{fig:vis_scannet} (1)), all matchers give similar numbers of inliers. However, both our IMP and EIMP obtain smaller rotation and translation errors than SuperGlue* and SGMNet because the embedded geometric information in our matching module. In the iteration process, rather than finding inliers from a cluster, both IMP and EIMP expand the areas with inliers, which allows our models to find more potential inliers and make the pose estimation more stable. When testing images become more difficult (Fig.~\ref{fig:vis_scannet} (2) and (3)), the behavior of our models in expanding inliers over the whole meaningful regions of images can be observed more clearly. Especially for Fig.~\ref{fig:vis_scannet} (3) where all keypoints are extracted from regions with repetitive textures, matching methods based on pure descriptors can hardly discriminate these keypoints, so SuperGlue* and SGMNet fail to give enough inliers. At this time, geometric constraints play an indispensable role at finding correct matches, which explains the success of IMP and EIMP.

%However, both SuperGlue* and SGMNet focus mainly on extracting inliers from clusters where keypoints are more discriminative rather than more geometric, resulting fewer inliers and larger pose errors.

\subsection{Qualitative results on YFCC100m}
\label{sec:rec:yfcc}

\begin{figure*}[t]
	\centering
	\includegraphics[width=.95\linewidth]{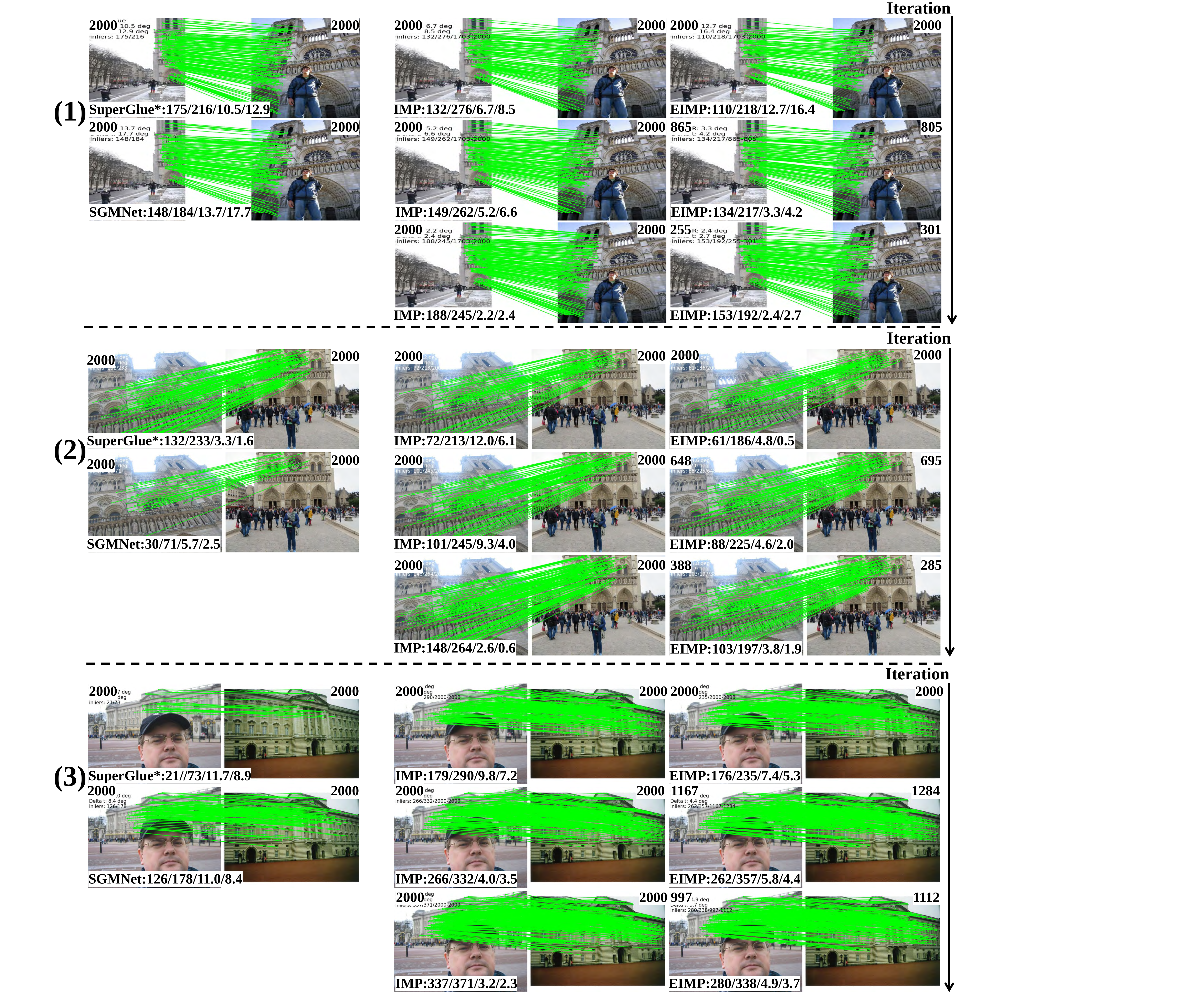}
	\caption{\textbf{Qualitative results on YFCC100m dataset~\cite{yfcc100m}} At the left-bottom of each image pair, we report the number of inliers/matches and rotation/translation errors of SuperGlue*~\cite{superglue} (official SuperGlue), SGMNet~\cite{sgmnet}, our IMP and EIMP. Besides, the number of keypoints in each image are shown at the top of each pair. For simple case (1), all methods give a large number of inliers, but IMP and EIMP report smaller pose errors even when using fewer inliers. Note that instead of keeping all keypoints, EIMP effectively reduces the number of keypoints from 2,0000 to 865 and 255 in the iteration process, significantly decreasing the time complexity for self and cross attention computation~\ref{fig:architecture}. For tough cases ((2), (3)), due to large viewpoint and illumination changes, SuperGlue* and SGMNet fail to report comparable number of inliers to our models, resulting in higher rotation and translation errors. In contrast, our models still progressively increase the number of inliers from different regions in the iteration process. These well distributed inliers lead to smaller pose errors. By comparing the results of EIMP in (2) and (3), we see that the number of preserved keypoints are based on the number of potential inliers in the image pair: more potential inliers result in more retrained keypoints. That is because our sampling strategy is fully adaptive.}
	\label{fig:vis_yfcc}
\end{figure*}

Fig.~\ref{fig:vis_yfcc} shows the predicted matches and relative poses reported by SuperGlue*~\cite{superglue}, SGMNet~\cite{sgmnet}, our IMP and EIMP. For simple cases (Fig.~\ref{fig:vis_yfcc} (1)), all methods give a large number of inliers while IMP and EIMP report smaller pose errors in the iteration process even using fewer inliers. Note that instead of keeping all keypoints, our EIMP with adaptive sampling effectively reduces the number of keypoints from 2,0000 to 865 and 255 in the iteration process, which significantly decreases the time complexity for self and cross attention computation~\ref{fig:architecture}. We also observe that for tough cases (Fig.~\ref{fig:vis_yfcc} (2) and (3)), due to large viewpoint and illumination changes, SuperGlue* and SGMNet fail to report comparable number of inliers to our models, resulting in higher rotation and translation errors. In contrast, our models still progressively increase the number of inliers from different regions in the iteration process. These well distributed inliers lead to smaller pose errors. By comparing the results of EIMP in Fig.~\ref{fig:vis_yfcc} (2) and Fig.~\ref{fig:vis_yfcc} (3), we see that the number of preserved keypoints are based on the number of potential inliers in the image pair: more potential inliers result in more retrained keypoints. That is because our sampling strategy is fully adaptive.

\subsection{Qualitative results on Aachen}
\label{sec:rec:aachen}

\begin{figure*}[t]
	\centering
	\includegraphics[width=.99\linewidth]{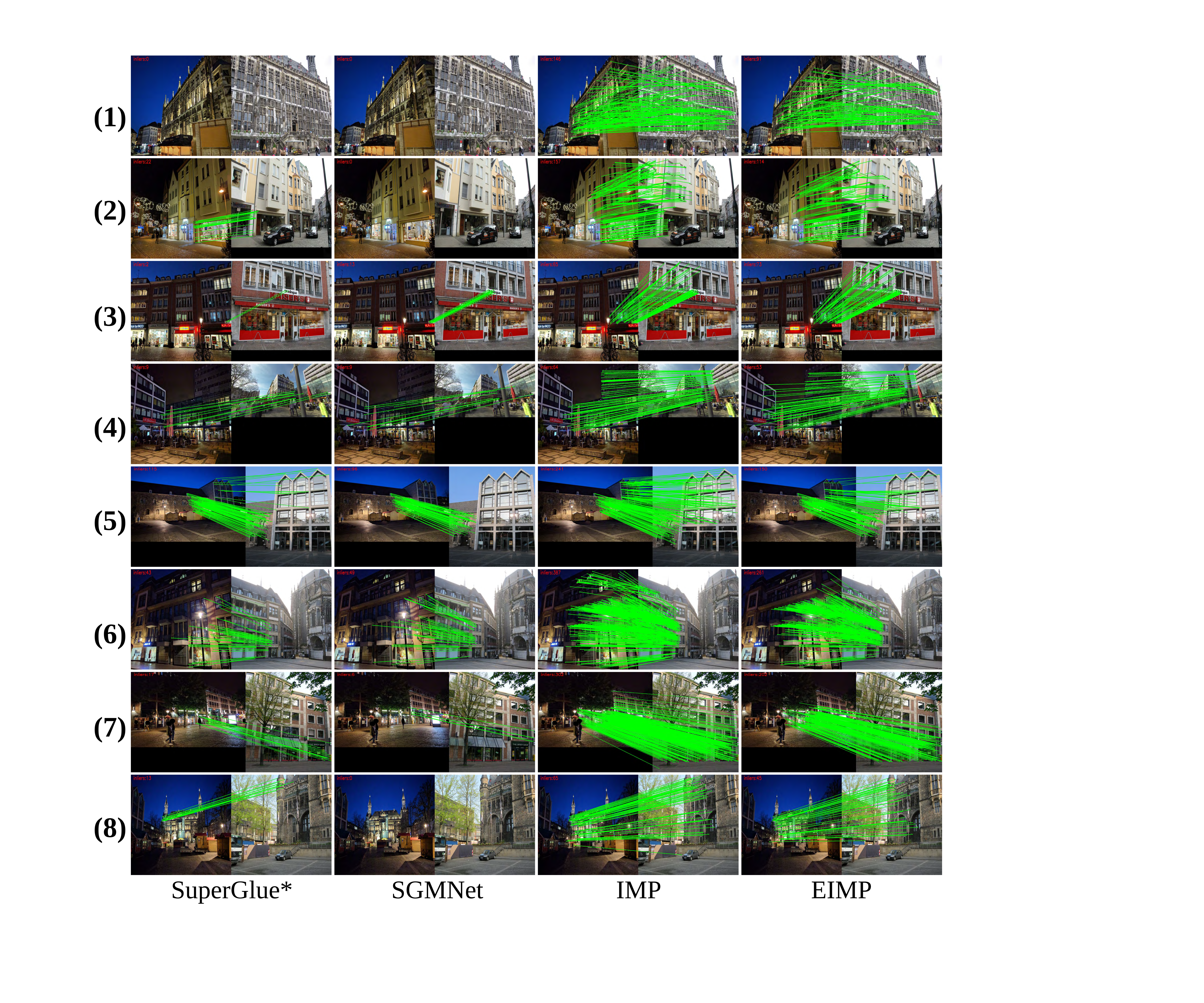}
	\caption{\textbf{Qualitative results on Aachen v1.1 dataset~\cite{aachen,aachenv112021}}. We visualize inliers of query (left) and reference (right) images under larger viewpoint ((1)-(4)) and illumination ((5)-(8)) changes of SuperGlue*~\cite{superglue} (official SuperGlue), SGMNet~\cite{sgmnet}, our IMP and EIMP. As the groundtruth poses of each query images are not available, we utilize HLoc~\cite{hloc2019} framework to visualize inliers given by PnP~\cite{epnp2009} + RANSAC~\cite{ransac}. Testing pairs (1)-(4) show that when image pairs have large viewpoint changes, both SuperGlue* and SGMNet fail to find enough correct matches. That is because geometric constraints are more useful for finding matches in two images with large viewpoint changes and both SuperGlue* and SGMNet ignore it. However, we embed the geometric information into the matching module, so our IMP and EIMP work much better, guaranteeing the localization success.  For cases (5)-(8), when query images have large illumination changes with reference images, corresponding keypoints from two images are less discriminative, so SuperGlue* and SGMNet only give slightly more inliers than for images with large viewpoint. As our model additionally leverages geometric constraints to find matches, both IMP and EIMP successfully obtain a large number of inliers. Note that compared to SuperGlue* and SGMNet, both IMP and EIMP find inliers from the almost the whole overlap regions of the two images as opposed to some clusters ((2)-(5), (8)). }
	\label{fig:vis_aachenv11}
\end{figure*}

In Fig.~\ref{fig:vis_aachenv11}, we show the inliers between query and reference images in the large-scale localization task on Aachen v1.1 dataset~\cite{aachen,aachenv112021}. Different with image pairs in Scannet~\cite{scannet} and YFCC100m~\cite{yfcc100m} datasets, query and reference images in Aachen dataset are captured under totally different conditions usually with extremely large viewpoint (Fig.~\ref{fig:vis_aachenv11} (1)-(4)) and illumination (Fig.~\ref{fig:vis_aachenv11} (5)-(8)) changes, making finding matches difficult. As the groundtruth poses of query images are not available, we use HLoc~\cite{hloc2019} framework to visualize the inliers between query and reference images after the PnP~\cite{epnp2009} + RANSAC~\cite{ransac} for all methods.

Fig.~\ref{fig:vis_aachenv11} ((1)-(4)) shows that when image pairs have large viewpoint changes, both SuperGlue* and SGMNet fail to find enough correct matches. That is because geometric constraints are more useful for finding matches in two images with large viewpoint changes and both SuperGlue* and SGMNet ignore this information. However, we embed the geometric information into the matching module, so our IMP and EIMP work much better, guaranteeing the localization success.

When query images have large illumination changes with reference images, corresponding keypoints from two images are less discriminative, so SuperGlue* and SGMNet only give slightly more inliers than for images with large viewpoint, as shown in Fig.~\ref{fig:vis_aachenv11}((5)-(8)). As our model additionally leverages geometric constraints to find matches, both IMP and EIMP successfully obtain a large number of inliers. Note that compared to SuperGlue* and SGMNet, both IMP and EIMP find inliers from the almost the whole overlap regions of the two images as opposed to some clusters (Fig.~\ref{fig:vis_aachenv11} (2)-(5) and (8)). 

%%%%%%%%% REFERENCES
{\small
\bibliographystyle{ieee_fullname}
\bibliography{egbib}

\begin{thebibliography}{10}\itemsep=-1pt

\bibitem{dinofeat2022}
Shir Amir, Yossi Gandelsman, Shai Bagon, and Tali Dekel.
\newblock {Deep vit features as dense visual descriptors}.
\newblock In {\em ECCV}, 2022.

\bibitem{netvald2016}
Relja Arandjelovic, Petr Gronat, Akihiko Torii, Tomas Pajdla, and Josef Sivic.
\newblock {NetVLAD: CNN architecture for weakly supervised place recognition}.
\newblock In {\em CVPR}, 2016.

\bibitem{rootsift2012}
Relja Arandjelovi{\'c} and Andrew Zisserman.
\newblock {Three things everyone should know to improve object retrieval}.
\newblock In {\em CVPR}, 2012.

\bibitem{probdepth2022}
Gwangbin Bae, Ignas Budvytis, and Roberto Cipolla.
\newblock {Multi-View Depth Estimation by Fusing Single-View Depth Probability
  with Multi-View Geometry}.
\newblock In {\em CVPR}, 2022.

\bibitem{goodmodel}
Daniel Barath, Luca Cavalli, and Marc Pollefeys.
\newblock {Learning To Find Good Models in RANSAC}.
\newblock In {\em CVPR}, 2022.

\bibitem{relativepose2022}
Daniel Barath and Zuzana Kukelova.
\newblock {Relative Pose from SIFT Features}.
\newblock In {\em ECCV}, 2022.

\bibitem{magsac}
Daniel Barath, Jiri Matas, and Jana Noskova.
\newblock {MAGSAC: marginalizing sample consensus}.
\newblock In {\em CVPR}, 2019.

\bibitem{magsac++}
Daniel Barath, Jana Noskova, Maksym Ivashechkin, and Jiri Matas.
\newblock {MAGSAC++, a fast, reliable and accurate robust estimator}.
\newblock In {\em CVPR}, 2020.

\bibitem{ngransac}
Eric Brachmann and Carsten Rother.
\newblock {N}eural-{G}uided {RANSAC}: {L}earning where to sample model
  hypotheses.
\newblock In {\em ICCV}, 2019.

\bibitem{dino2021}
Mathilde Caron, Hugo Touvron, Ishan Misra, Herv{\'e} J{\'e}gou, Julien Mairal,
  Piotr Bojanowski, and Armand Joulin.
\newblock {Emerging properties in self-supervised vision transformers}.
\newblock In {\em ICCV}, 2021.

\bibitem{adalam}
Luca Cavalli, Viktor Larsson, Martin~Ralf Oswald, Torsten Sattler, and Marc
  Pollefeys.
\newblock {Handcrafted outlier detection revisited}.
\newblock In {\em ECCV}, 2020.

\bibitem{shareattention}
Boyu Chen, Peixia Li, Baopu Li, Chuming Li, Lei Bai, Chen Lin, Ming Sun, Junjie
  Yan, and Wanli Ouyang.
\newblock {PSViT: Better vision transformer via token pooling and attention
  sharing}.
\newblock {\em arXiv preprint arXiv:2108.03428}, 2021.

\bibitem{sgmnet}
Hongkai Chen, Zixin Luo, Jiahui Zhang, Lei Zhou, Xuyang Bai, Zeyu Hu, Chiew-Lan
  Tai, and Long Quan.
\newblock {Learning to match features with seeded graph matching network}.
\newblock In {\em CVPR}, 2021.

\bibitem{degensac}
Ondrej Chum, Tomas Werner, and Jiri Matas.
\newblock {Two-view geometry estimation unaffected by a dominant plane}.
\newblock In {\em CVPR}, 2005.

\bibitem{sinkhorn2013}
Marco Cuturi.
\newblock {Sinkhorn distances: Lightspeed computation of optimal transport}.
\newblock In {\em NIPS}, 2013.

\bibitem{scannet}
Angela Dai, Matthias Nie{\ss}ner, Michael Zollh{\"o}fer, Shahram Izadi, and
  Christian Theobalt.
\newblock {Bundlefusion: Real-time globally consistent 3d reconstruction using
  on-the-fly surface reintegration}.
\newblock {\em ACM ToG}, 2017.

\bibitem{ms2dg}
Luanyuan Dai, Yizhang Liu, Jiayi Ma, Lifang Wei, Taotao Lai, Changcai Yang, and
  Riqing Chen.
\newblock {MS2DG-Net: Progressive Correspondence Learning via Multiple Sparse
  Semantics Dynamic Graph}.
\newblock In {\em CVPR}, 2022.

\bibitem{superpoint}
Daniel DeTone, Tomasz Malisiewicz, and Andrew Rabinovich.
\newblock {Superpoint: Self-supervised interest point detection and
  description}.
\newblock In {\em CVPRW}, 2018.

\bibitem{instability}
Hongyi Fan, Joe Kileel, and Benjamin Kimia.
\newblock {On the Instability of Relative Pose Estimation and RANSAC's Role}.
\newblock In {\em CVPR}, 2022.

\bibitem{adavit2022}
Mohsen Fayyaz, Soroush~Abbasi Koohpayegani, Farnoush Rezaei, and Sommerlade1
  Hamed Pirsiavash2~Juergen Gall.
\newblock {Adaptive Token Sampling For Efficient Vision Transformers}.
\newblock In {\em ECCV}, 2022.

\bibitem{ransac}
Martin~A Fischler and Robert~C Bolles.
\newblock {Random sample consensus: a paradigm for model fitting with
  applications to image analysis and automated cartography}.
\newblock {\em Communications of the ACM}, 1981.

\bibitem{mvg}
Richard Hartley and Andrew Zisserman.
\newblock {\em {Multiple view geometry in computer vision}}.
\newblock Cambridge university press, 2003.

\bibitem{linearattention}
Angelos Katharopoulos, Apoorv Vyas, Nikolaos Pappas, and Fran{\c{c}}ois
  Fleuret.
\newblock {Transformers are rnns: Fast autoregressive transformers with linear
  attention}.
\newblock In {\em ICML}, 2020.

\bibitem{adam}
Diederik~P Kingma and Jimmy Ba.
\newblock {Adam: A method for stochastic optimization}.
\newblock In {\em ICLR}, 2015.

\bibitem{setformer2019}
Juho Lee, Yoonho Lee, Jungtaek Kim, Adam Kosiorek, Seungjin Choi, and Yee~Whye
  Teh.
\newblock {Set transformer: A framework for attention-based
  permutation-invariant neural networks}.
\newblock In {\em ICML}, 2019.

\bibitem{epnp2009}
Vincent Lepetit, Francesc Moreno-Noguer, and Pascal Fua.
\newblock {EPnP: An Accurate O(n) Solution to the PnP Problem}.
\newblock {\em IJCV}, 81:155--166, 2009.

\bibitem{megadepth}
Zhengqi Li and Noah Snavely.
\newblock {Megadepth: Learning single-view depth prediction from internet
  photos}.
\newblock In {\em CVPR}, 2018.

\bibitem{pixelperfect}
Philipp Lindenberger, Paul-Edouard Sarlin, Viktor Larsson, and Marc Pollefeys.
\newblock {Pixel-Perfect Structure-from-Motion with Featuremetric Refinement}.
\newblock In {\em ICCV}, 2021.

\bibitem{lmcnet}
Yuan Liu, Lingjie Liu, Cheng Lin, Zhen Dong, and Wenping Wang.
\newblock {Learnable motion coherence for correspondence pruning}.
\newblock In {\em CVPR}, 2021.

\bibitem{sift}
David~G Lowe.
\newblock {Distinctive image features from scale-invariant keypoints}.
\newblock {\em IJCV}, 2004.

\bibitem{pytorch}
Adam Paszke, Sam Gross, Francisco Massa, Adam Lerer, James Bradbury, Gregory
  Chanan, Trevor Killeen, Zeming Lin, Natalia Gimelshein, Luca Antiga, et~al.
\newblock {Pytorch: An imperative style, high-performance deep learning
  library}.
\newblock In {\em NeurIPS}, 2019.

\bibitem{transport2017}
Gabriel Peyr{\'e} and Marco Cuturi.
\newblock {Computational optimal transport}.
\newblock {\em Foundations and Trends{\textregistered} in Machine Learning},
  11(5-6):355--607, 2019.

\bibitem{oxford2018}
Filip Radenovi{\'c}, Ahmet Iscen, Giorgos Tolias, Yannis Avrithis, and
  Ond{\v{r}}ej Chum.
\newblock {Revisiting oxford and paris: Large-scale image retrieval
  benchmarking}.
\newblock In {\em CVPR}, 2018.

\bibitem{usac}
Rahul Raguram, Ondrej Chum, Marc Pollefeys, Jiri Matas, and Jan-Michael Frahm.
\newblock {USAC: A universal framework for random sample consensus}.
\newblock {\em TPAMI}, 2012.

\bibitem{dynamicvit}
Yongming Rao, Wenliang Zhao, Benlin Liu, Jiwen Lu, Jie Zhou, and Cho-Jui Hsieh.
\newblock {Dynamicvit: Efficient vision transformers with dynamic token
  sparsification}.
\newblock In {\em NeurIPS}, 2021.

\bibitem{orb}
Ethan Rublee, Vincent Rabaud, Kurt Konolige, and Gary Bradski.
\newblock {ORB: An efficient alternative to SIFT or SURF}.
\newblock In {\em ICCV}, 2011.

\bibitem{hloc2019}
Paul-Edouard Sarlin, Cesar Cadena, Roland Siegwart, and Marcin Dymczyk.
\newblock {From Coarse to Fine: Robust Hierarchical Localization at Large
  Scale}.
\newblock In {\em CVPR}, 2019.

\bibitem{superglue}
Paul-Edouard Sarlin, Daniel DeTone, Tomasz Malisiewicz, and Andrew Rabinovich.
\newblock {Superglue: Learning feature matching with graph neural networks}.
\newblock In {\em CVPR}, 2020.

\bibitem{aachen}
Torsten Sattler, Will Maddern, Carl Toft, Akihiko Torii, Lars Hammarstrand,
  Erik Stenborg, Daniel Safari, Masatoshi Okutomi, Marc Pollefeys, Josef Sivic,
  et~al.
\newblock {Benchmarking 6dof outdoor visual localization in changing
  conditions}.
\newblock In {\em CVPR}, 2018.

\bibitem{clustergnn}
Yan Shi, Jun-Xiong Cai, Yoli Shavit, Tai-Jiang Mu, Wensen Feng, and Kai Zhang.
\newblock {ClusterGNN: Cluster-based Coarse-to-Fine Graph Neural Network for
  Efficient Feature Matching}.
\newblock In {\em CVPR}, 2022.

\bibitem{sinkhorn1967}
Richard Sinkhorn and Paul Knopp.
\newblock {Concerning nonnegative matrices and doubly stochastic matrices}.
\newblock {\em Pacific Journal of Mathematics}, 21(2):343--348, 1967.

\bibitem{acne}
Weiwei Sun, Wei Jiang, Eduard Trulls, Andrea Tagliasacchi, and Kwang~Moo Yi.
\newblock {Acne: Attentive context normalization for robust
  permutation-equivariant learning}.
\newblock In {\em CVPR}, 2020.

\bibitem{ela2022}
Suwichaya Suwanwimolkul and Satoshi Komorita.
\newblock Efficient linear attention for fast and accurate keypoint matching.
\newblock In {\em ICMR}, 2022.

\bibitem{efficient_transformer}
Yi Tay, Mostafa Dehghani, Dara Bahri, and Donald Metzler.
\newblock {Efficient transformers: A survey}.
\newblock {\em ACM Computing Surveys}, 2020.

\bibitem{yfcc100m}
Bart Thomee, David~A Shamma, Gerald Friedland, Benjamin Elizalde, Karl Ni,
  Douglas Poland, Damian Borth, and Li-Jia Li.
\newblock {YFCC100M: The new data in multimedia research}.
\newblock {\em Communications of the ACM}, 2016.

\bibitem{deit}
Hugo Touvron, Matthieu Cord, Matthijs Douze, Francisco Massa, Alexandre
  Sablayrolles, and Herv{\'e} J{\'e}gou.
\newblock {Training data-efficient image transformers \& distillation through
  attention}.
\newblock In {\em ICML}, 2021.

\bibitem{attention}
Ashish Vaswani, Noam Shazeer, Niki Parmar, Jakob Uszkoreit, Llion Jones,
  Aidan~N Gomez, {\L}ukasz Kaiser, and Illia Polosukhin.
\newblock {Attention is all you need}.
\newblock In {\em NeurIPS}, 2017.

\bibitem{linformer2020}
Sinong Wang, Belinda~Z Li, Madian Khabsa, Han Fang, and Hao Ma.
\newblock {Linformer: Self-attention with linear complexity}.
\newblock {\em arXiv preprint arXiv:2006.04768}, 2020.

\bibitem{sun3d2013}
Jianxiong Xiao, Andrew Owens, and Antonio Torralba.
\newblock {Sun3d: A database of big spaces reconstructed using sfm and object
  labels}.
\newblock In {\em ICCV}, 2013.

\bibitem{sfd2}
Fei Xue, Ignas Budvytis, and Roberto Cipolla.
\newblock {SFD2: Semantic-guided Feature Detection and Description}.
\newblock In {\em CVPR}, 2023.

\bibitem{lsg}
Fei Xue, Xin Wang, Zike Yan, Qiuyuan Wang, Junqiu Wang, and Hongbin Zha.
\newblock Local supports global: Deep camera relocalization with sequence
  enhancement.
\newblock In {\em ICCV}, 2019.

\bibitem{glnet}
Fei Xue, Xin Wu, Shaojun Cai, and Junqiu Wang.
\newblock {Learning multi-view camera relocalization with graph neural
  networks}.
\newblock In {\em CVPR}, 2020.

\bibitem{ce}
Kwang~Moo Yi, Eduard Trulls, Yuki Ono, Vincent Lepetit, Mathieu Salzmann, and
  Pascal Fua.
\newblock {Learning to find good correspondences}.
\newblock In {\em CVPR}, 2018.

\bibitem{oanet}
Jiahui Zhang, Dawei Sun, Zixin Luo, Anbang Yao, Lei Zhou, Tianwei Shen, Yurong
  Chen, Long Quan, and Hongen Liao.
\newblock {Learning two-view correspondences and geometry using order-aware
  network}.
\newblock In {\em ICCV}, 2019.

\bibitem{aachenv112021}
Zichao Zhang, Torsten Sattler, and Davide Scaramuzza.
\newblock {Reference pose generation for long-term visual localization via
  learned features and view synthesis}.
\newblock {\em IJCV}, 2021.

\bibitem{clnet}
Chen Zhao, Yixiao Ge, Feng Zhu, Rui Zhao, Hongsheng Li, and Mathieu Salzmann.
\newblock {Progressive correspondence pruning by consensus learning}.
\newblock In {\em ICCV}, 2021.

\bibitem{tnet}
Zhen Zhong, Guobao Xiao, Linxin Zheng, Yan Lu, and Jiayi Ma.
\newblock {T-Net: Effective Permutation-Equivariant Network for Two-View
  Correspondence Learning}.
\newblock In {\em ICCV}, 2021.

\end{thebibliography}
}

\end{document}